\documentclass[manuscript,screen]{acmart}

\AtBeginDocument{%
  \providecommand\BibTeX{{%
    \normalfont B\kern-0.5em{\scshape i\kern-0.25em b}\kern-0.8em\TeX}}}
    
\usepackage{graphicx}
\usepackage{blindtext}
\usepackage{xpatch}
\makeatletter
\let\@authorsaddresses\@empty
\makeatother

\setcopyright{none}
\renewcommand\footnotetextcopyrightpermission[1]{}
\usepackage{graphicx}
\usepackage{algorithm}

\usepackage[inline]{enumitem}
\usepackage{algorithmic}
\usepackage{graphicx}
\usepackage{caption}
\usepackage{subcaption}

\newcommand{\eat}[1]{}

\begin{document}

\title{RFID-Based Indoor Spatial Query Evaluation with Bayesian Filtering Techniques}

\author{Bo Hui}
\affiliation{%
 \institution{Auburn University}}
 
\author{Wenlu Wang}
\affiliation{%
 \institution{TAMUCC}}
 
 \author{Jiao Yu}
\affiliation{%
 \institution{Auburn University}}
 
 \author{Zhitao Gong}
\affiliation{%
 \institution{Auburn University}}
 
 \author{Wei-Shinn~Ku}
\affiliation{%
 \institution{Auburn University}}

 \author{Min-Te Sun}
\affiliation{%
 \institution{National Central University}
}

 \author{Hua Lu}
\affiliation{%
 \institution{Roskilde University}
}

\begin{abstract}
People spend a significant amount of time in indoor spaces (e.g.,
office buildings, subway systems, etc.) in their daily lives.
Therefore, it is important to develop efficient indoor spatial
query algorithms for supporting various location-based
applications. However, indoor spaces differ from outdoor spaces
because users have to follow the indoor floor plan for their
movements. In addition, positioning in indoor environments is
mainly based on sensing devices (e.g., RFID readers) rather than
GPS devices. Consequently, we cannot apply existing spatial query
evaluation techniques devised for outdoor environments for this
new challenge. Because Bayesian filtering techniques can be
employed to estimate the state of a system that changes over time
using a sequence of noisy measurements made on the system, in this
research, we propose the Bayesian filtering-based location
inference methods as the basis for evaluating indoor spatial
queries with noisy RFID raw data. Furthermore, two novel models,
indoor walking graph model and anchor point indexing model, are
created for tracking object locations in indoor environments.
Based on the inference method and tracking models, we develop
innovative indoor range and $k$ nearest neighbor ($k$NN) query
algorithms. We validate our solution through use of both synthetic data and real-world data. Our experimental results show that the
proposed algorithms can evaluate indoor spatial queries
effectively and efficiently. We open-source the code, data, and floor plan at \color{blue}https://github.com/DataScienceLab18/IndoorToolKit.

\end{abstract}
\keywords{Indoor spatial query, RFID, Bayesian filtering}
\maketitle
\thispagestyle{empty}

\section{Introduction}\label{sec:intro}
Today most people spend a significant portion of their time daily
in indoor spaces such as subway systems, office buildings,
shopping malls, convention centers, and many other structures. In
addition, indoor spaces are becoming increasingly large and
complex. For instance, the New York City Subway has 472 stations
and contains 245 miles (394 km) of routes~\cite{NYCS}. In 2017,
the subway system delivered over 1.73 billion rides, averaging
approximately 5.6 million rides on weekdays~\cite{MTA}. Therefore,
users will have more and more demand for launching spatial queries
for finding friends or Points Of Interest (POI)~\cite{wang2017efficient,wang2019scalable} in indoor places.
Moreover, users are usually moving around when issuing queries.
Thus we need to properly support indoor spatial queries
\emph{continuously}, e.g., reporting nearby friends in a mall when
a user is shopping. However, existing spatial query evaluation
techniques for outdoor environments (either based on Euclidean
distance or network
distance)~\cite{conf/sigmod/RoussopoulosKV95,journals/tods/HjaltasonS99,conf/vldb/PapadiasZMT03,conf/sigmod/SametSA08,journals/tkde/LeeLZT12}
cannot be applied in indoor spaces because these techniques assume
that user locations can be acquired from GPS signals or cellular
positioning, but the assumption does not hold in covered indoor
spaces. Furthermore, indoor spaces are usually modelled
differently from outdoor spaces. In indoor environments, user
movements are enabled or constrained by topologies such as doors, walls, and hallways.

Radio Frequency Identification (RFID) technologies have become
increasingly popular over the last decade with applications in
areas such as supply chain
management~\cite{journals/cacm/SantosS08}, health care~\cite{amendola2014rfid}, and
transportation~\cite{jedermann2009spatial}. In indoor environments, RFID is mainly employed to
support track and trace applications. Generally, RFID readers are
deployed in critical locations while objects carry RFID tags. When
a tag passes the detection range of a reader, the reader
recognizes the presence of the tag and generates a record in the
back end database. However, the raw data collected by RFID readers
is inherently
unreliable~\cite{Sullivan0504,conf/vldb/JefferyGF06}, with false
negatives as a result of RF interference, limited detection range,
tag orientation, and other environmental
phenomena~\cite{conf/mobisys/WelbourneKSBB09}. In addition,
readers cannot cover all areas of interest because of their high
cost or privacy
concerns~\cite{journals/internet/WelbourneBCGRRBB09}. Therefore,
we cannot directly utilize RFID raw data to evaluate commonly used
spatial query types (e.g., range and $k$NN) for achieving high
accuracy results in indoor environments. In addition, several
other types of wireless communication technologies such as Wi-Fi
and Bluetooth have been employed for indoor
positioning~\cite{conf/icdcsw/AnastasiBCDGM03,conf/gis/BellJK10}.
However, each aforementioned technology has considerable
positioning uncertainty. Furthermore, Wi-Fi and Bluetooth are
mainly utilized for positioning individual users rather than
supporting a centralized indoor location tracking system. It is
too expensive to attach Wi-Fi or Bluetooth transmitters (\$5 per
device) to monitored objects. Therefore, we focus on RFID in this
research.

In this paper, we consider the setting of an indoor environment
where a number of RFID readers are deployed in hallways. Each user
is attached with an RFID tag, which can be identified by a reader
when the user is within the detection range of the reader. Given
the history of RFID raw readings from all the readers, we are in
a position to design a system that can efficiently answer indoor
spatial queries. We mainly focus on four types of spatial queries,
range query, $k$NN query, continuous range query, and continuous
$k$NN query.

%
%
%


Bayesian filtering
techniques~\cite{Arulampalama:tutorial,Maybeck79} can be employed
to estimate the state of a system that changes over time using a
sequence of noisy measurements made on the system. In this paper
we propose the Bayesian filtering-based location inference
methods, the indoor walking graph model, and the anchor point
indexing model for inferring object locations from noisy RFID raw
data. On top of the location inference, indoor spatial queries can
be evaluated efficiently by our algorithms with high accuracy. The
contributions of this study are as follows:

\begin{itemize}
\item We design the Bayesian filtering-based location inference
methods as the basis for evaluating indoor spatial queries.

\item We propose two novel models, the indoor walking graph model
and the anchor point indexing model, and an RFID-based system for
tracking object locations in indoor environments.

\item Indoor spatial query evaluation algorithms for range, $k$NN,
continuous range, and continuous $k$NN queries are developed based
on the proposed system.

\item We demonstrate the efficiency and effectiveness of our
approach by comparing the performance of our system with the
symbolic model-based solution~\cite{Yang:indoorknn} through
experiments based on real-world data and synthetic data.
\end{itemize}

The rest of this paper is organized as follows. In
Section~\ref{sec:related}, we survey previous works for indoor
object monitoring and spatial queries. Background knowledge of
particle filters and the Kalman filter is provided in
Section~\ref{sec:prelim}. In Section~\ref{sec:design}, we
introduce our Bayesian filtering-based indoor spatial query
evaluation system. The experimental validation of our design is
presented in Section~\ref{sec:expr}. Section~\ref{sec:conc}
concludes this paper with a discussion of future work.

\section{Related Work}\label{sec:related}

In this section, we review previous work related to indoor spatial
queries and RFID data cleansing.

\subsection{Indoor Spatial Queries}

Outdoor spatial queries, e.g., range and $k$NN queries, have been
extensively studied both for Euclidean
space~\cite{conf/sigmod/RoussopoulosKV95,journals/tods/HjaltasonS99}
and road
networks~\cite{conf/vldb/PapadiasZMT03,conf/sigmod/SametSA08,
journals/tkde/LeeLZT12}. However, due to the inherent differences
in spatial characteristics, indoor spatial queries need different
models and cannot directly apply mature techniques from their
outdoor counterparts. Therefore, indoor spatial queries are
drawing more and more research attention from industry and
academia. For answering continuous range queries in indoor
environments, Jensen \emph{et
al.}~\cite{Jensen:2009:GMB:1590953.1591000} proposed using the
\emph{positioning device deployment graph} to represent the
connectivity of rooms and hallways from the perspective of
positioning devices. Basically, entities that can be accessed
without having to be detected by any positioning device are
represented by one cell in the graph, and edges connecting two
cells in the graph represent the positioning device(s) which
separate them. Based on the graph, initial query results can be
easily processed with the help of an indexing scheme also proposed
by the authors~\cite{Yang:indoorrange}. Query results are returned
in two forms: certain results and uncertain results. To reduce the
workload of maintaining and updating the query results, Yang
\emph{et al.} further proposed the concept of \emph{critical
devices}. Only from the ENTER and LEAVE observations of its
critical devices can a query's results be affected. However, the
probability model utilized in Yang's work is very simple: a moving
object is uniformly distributed over all the reachable locations
constrained by its maximum speed in a given indoor space. This
simple probability model is incapable of taking advantage of the
moving object's previous moving patterns, such as direction and
speed, which would make the location prediction more reasonable
and precise. In addition, Yang \emph{et al.}~\cite{Yang:indoorknn}
also addressed the problem of $k$NN queries over moving objects in
indoor spaces. Unlike another previous
work~\cite{DBLP:dblp_conf/mdm/LiL08} which defines nearest
neighbors by the minimal number of doors to go through, they
proposed a novel distance metric, minimum indoor walking distance,
as the underlying metric for indoor $k$NN queries. Moreover, Yang
\emph{et al.} provided the formal definition for Indoor
Probabilistic Threshold $k$NN Query (PT$k$NN) as finding a result
set with $k$ objects which have a higher probability than the
threshold probability $T$. Indoor distance-based pruning and
probability threshold-based pruning are proposed in Yang's work to
speed up PT$k$NN query processing. Similarly, the paper employs
the same simple probabilistic model as in~\cite{Yang:indoorrange},
and, therefore, has the same deficiencies in probability
evaluation.
An adaptive cleansing (AC) probabilistic model~\cite{zhao2012model} is proposed
to achieve object tracking in open spaces.
An RFID data cleaning method that optimizes the overall accuracy and cost is 
proposed in~\cite{gonzalez2007cost}. 
However, ~\cite{zhao2012model} and~\cite{gonzalez2007cost} are 
different from our event-driven setting, because indoor topology is not applied.
A spatial cleansing model~\cite{baba2013spatiotemporal} that utilizes a distance-aware graph to reduce spatial ambiguity in indoor spaces is proposed for RFID data cleansing. 
Their method is more focused on predicting the actual location among 
its alternative possibilities rather than solving spatial queries. 
Offline cleaning with sub-sequence data~\cite{fazzinga2014offline} is also taken into consideration. 
Their method is applicable only when data are stabilized and used for analysis tasks.
The main contribution of~\cite{fazzinga2014cleaning}
is a framework which cleans RFID data by utilizing reachability and travel time limits.
~\cite{fazzinga2014offline} and~\cite{fazzinga2014cleaning} 
suffer from certain constraints and can not be applied to on-line spatial queries. To employ different methods in different user scenarios, \cite{msn-indoor} uses a pre-trained Neural Network model to classify users into different categories.

\subsection{RFID-Based Track and Trace}

RFID is a very popular electronic tagging technology that allows
objects to be automatically identified at a distance using an
electromagnetic challenge-and-response exchange of
data~\cite{journals/queue/Want04}. An RFID-based system consists
of a large number of low-cost tags that are attached to objects,
and readers which can identify tags without a direct line-of-sight
through RF communications. RFID technologies enable exceptional
visibility to support numerous track and trace applications in
different fields~\cite{conf/percom/YangCZT12},
including indoor navigation~\cite{wang1,wang2} and indoor trajectory mining~\cite{mine1,mine2}. However, the raw
data collected by RFID readers is inherently noisy and
inconsistent~\cite{Sullivan0504,conf/vldb/JefferyGF06}. Therefore,
middleware systems are required to correct readings and provide
cleansed data~\cite{journals/vldb/JefferyFG08}. In addition to the
unreliable nature of RFID data streams, another limitation is that
due to the high cost of RFID readers, RFID readers are mostly
deployed such that they have disjointed activation ranges in the
settings of indoor tracking. 



To overcome the above limitations, RFID data cleansing is a
necessary step to produce consistent data to be utilized by
high-level applications. Baba \emph{et al.}~\cite{6916912}
proposed a probabilistic distance-aware graph model to handle
false negative in RFID readings. The main limitation is that their
generative model relies on a long tracking history to detect and
possibly correct RFID readings. Tran \emph{et
al.}~\cite{DBLP:yanlei} used a sampling-based method called
particle filtering to infer clean and precise event streams from
noisy raw data produced by mobile RFID readers. Three enhancements
are proposed in their work to make traditional particle filter
techniques scalable. However, their work is mainly designed for
warehouse settings where objects remain static on shelves, which
is quite different from our setting where objects move around in a
building. Therefore, Tran's approach of adapting and applying
particle filters cannot be directly applied to our settings.
Another limitation of~\cite{DBLP:yanlei} is that they did not
explore further utilization of the output event streams for
high-level applications. Chen \emph{et
  al.}~\cite{haiquan,journals/tkde/Ku12} employed a different sampling
method called Markov Chain Monte Carlo (MCMC) to infer objects'
locations on shelves in warehouses.  Their method takes advantage
of the spatial and temporal redundancy of raw RFID readings, and
also considers environmental constraints such as the capacity of
shelves, to make the sampling process more precise. Their work
also focuses on warehouse settings; thus it is not suitable for
our problem of general indoor settings. The works
in~\cite{conf/sigmod/ReLBS08,conf/mobisys/WelbourneKLLBBS08,
  conf/icde/LetchnerRBP09} target settings such as office buildings,
which are similar to our problem. They use particle filters in
their preprocessing module to generate probabilistic streams, on
which complex event queries such as ``Is Joe meeting with Mary in
Room 203?'' can be processed. However, their goal is to answer
event queries instead of spatial queries, which is different from
the goal of this research. Geng \emph{et al.}~\cite{6655909} also
proposed using particle filters for indoor tracing with RFID;
however, they assumed a grid layout of RFID readers instead of
only along the hallways. Thus their algorithms cannot be applied
to our problem.


\section{Preliminary}\label{sec:prelim}
In this section, we briefly introduce the mathematical background
of Bayesian filters, including the Kalman filter and particle
filters, and location inference based on the two filters.
Notations used in this paper are summarized in Table~I.
\begin{table}[!b]
\vspace{-3mm}
  \centering
  \begin{tabular}{|c|p{2.6in}|}
    \hline
    Symbol & Meaning \\
    \hline \hline
    $q$ & An indoor query point\\
    $o_i$ & The object with ID $i$\\
    $C$ & A set of candidate objects\\
    $D$ & A set of sensing devices\\
    $G$ & The indoor walking graph\\
    $E$ & The edge set of $G$\\
    $N$ & The node (i.e., intersection) set of $G$\\
    $p_i$ & A probability distribution function for $o_i$ in terms of all possible locations\\
    $ap_i$ & An anchor point with ID $i$\\
    $N_s$ & The total number of particles for an object \\
    $u_{max}$ & The maximum walking speed of a person\\
    $l_{max}$ & The maximum walking distance of a person during a certain period of time\\
    $UR(o_i)$ & The uncertain region of object $o_i$\\
    $s_i$ & The minimum shortest network distance\\
    $l_i$ & The maximum shortest network distance\\
    $Area_i$ & The size of a given region $i$\\
    \(d_i\) & The \(i\)th RFID reader\\
    $prob(a,t)$ &  The probability that object $a$ exists at the searched location at time $t$.\\
    $prob(A,q,t)$ & The total probability of all objects in the result set with query $q$ at time $t$ \\
    \hline
  \end{tabular}
  \caption{Symbolic Notations}\label{sntab}
\end{table}

\subsection{The Kalman Filter}\label{sec:KFprelim}

The Kalman filter is an optimal recursive data processing
algorithm, which combines a system's dynamics model, known control
inputs, and observed measurements to form an optimal estimate of
system states. Note here the control inputs and observed
measurements are not deterministic, but rather with a degree of uncertainty. The Kalman filter works by making a
prediction of the future system state, obtaining measurements for
that future state, and adjusting its estimate by moderating the
difference between the two. The result of the Kalman filter is a
new probability distribution of system state which has reduced its
uncertainty to be less than either the original predicted values
or measurements alone.

To help readers better understand how the Kalman filter works for
location estimation, we use a simple example of one dimensional
movement and location estimation. Suppose an object is moving
along a horizontal line, and we are interested in estimating the
object's location $x$ with the Kalman filter. We assume the
object's speed can be expressed by $d_x/d_t=u+w$, where $u$ is a
constant and $w$ is a Gaussian variable with a mean of zero and
variance of $\sigma_w^2$. We also assume the object's initial
location at $t_0$ is also a Gaussian distribution with mean
$\hat{x}_0$ and variance $\sigma^2_0$. At a later time $t_{1^-}$,
just before an observation is made, we get a prediction of the
object's location $x_{1^-}$ to be a Gaussian distribution with
mean and variance:

\begin{equation}
 \hat{x}_{1^-}=\hat{x}_0+u*(t_1-t_0)
\end{equation}

\begin{equation}\label{eq:predvar}
\sigma_{1^-}^2=\sigma_0^2+\sigma_w^2*(t_1-t_0)
\end{equation}

As indicated by Equation~\ref{eq:predvar}, the uncertainty in the
predicted location $x_1$ increases with the time span $t_1-t_0$,
since no measurements are made during the time span and the
uncertainty in speed accumulates with time.

After the observation at $t_1$ is made, suppose its value turns
out to be $z_1$ with variance $\sigma_{z_1}^2$. The Kalman filter
combines the predicted value with the measured value to yield an
optimal estimation with mean and variance:

\begin{equation}\label{eq:kalmanmean}
 \hat{x}_{1}=\hat{x}_{1^-}+K_1*(z_1-\hat{x}_{1^-})
\end{equation}

\begin{equation}\label{eq:kalmanvar}
\sigma_{1}^2=\sigma_{1^-}^2-K_1*\sigma_{1^-}^2
\end{equation}

where $K_1=\sigma_{1^-}^2/(\sigma_{1^-}^2+\sigma_{z_1}^2)$. The
details of deriving Equations~\ref{eq:kalmanmean}
and~\ref{eq:kalmanvar} are omitted here, and we refer readers
to~\cite{Maybeck79} for further details.

As we can see from Equation~\ref{eq:kalmanmean}, the optimal
estimate $\hat{x}_{1}$ is the optimal predicted value before the
measurement plus a correction term. The variance $\sigma_{1}^2$ is
smaller than either $\sigma_{1^-}^2$ or $\sigma_{z_1}^2$. The
optimal gain $K_1$ gives more weight to the better value (with
lower variance), so that if the prediction is more accurate than
the measurement, then $\hat{x}_{1^-}$ is weighed more; otherwise
$z_1$ is weighed more.

\subsection{The Particle Filter}
\label{sec:pf}


A particle filter is a method that can be applied to nonlinear
recursive Bayesian filtering
problems~\cite{Arulampalama:tutorial}. The system under
investigation is often modeled as a state vector $x_{k}$, which
contains all relevant information about the system at time $k$.
The observation $z_{k}$ at time $k$ is nonlinear to the true
system state $x_{k}$; also the system evolves from $x_{k}$ to
$x_{k+1}$ nonlinearly.

The objective of the particle filter method is to construct a
discrete approximation to the probability density function (pdf)
$p(x_k|z_{1:k})$ by a set of random samples with associated
weights. We denote the weight of the $i^{th}$ particle at time $k$
by $w_k^i$, and the $i^{th}$ particle at time $k$ by $x_k^i$. According to the mathematical equations of particle
filters~\cite{Arulampalama:tutorial}, the new weight $w_k^i$ is
proportional to the old weight $w_{k-1}^i$ augmented by the
observation likelihood $p(z_k|x_k^i)$. Thus, particles which are
more likely to cause an observation consistent with the true
observation result $z_k$ will gain higher weight than others.

The posterior filtered density $p(x_k|z_{1:k})$ can be
approximated as:

\begin{equation}
p(x_k|z_{1:k})\approx \sum_{i=1}^{N_s} w_k^i \delta(x_k-x_k^i)
\end{equation}
\begin{equation}
\delta (x)=\begin{cases}\infty, &x = 0 \cr 0, &x \neq 0\end{cases}
\end{equation}
\begin{equation}
\int_{-\infty}^{+\infty} \delta (x) dx= 1.
\end{equation}
Resampling is a method to solve the degeneration problem in
particle filters. Degeneration means that with more iterations
only a few particles would have dominant weights while the
majority of others would have near-zero weights. The basic idea of
resampling is to eliminate low weight particles, replicate high
weight particles, and generate a new set of particles
$\{x_k^{i_*}\}_{i=1}^{N_s}$ with equal weights. Our work adopts
sampling importance resampling filters, which perform the
resampling step at every time index.

In our application, particles update their locations according to
the object motion model employed in our work. Briefly, the object
motion model assumes objects move forward with constant speeds,
and can either enter rooms or continue to move along hallways.
Weights of particles are updated according to the device sensing
model~\cite{haiquan} used in this research. An example of applying
particle filters to the problem of RFID-based indoor location
inferences can be found in~\cite{conf/edbt/YuKSL13}.

\subsection{Query Definitions}\label{sec:def}
Here we define the probabilistic $k$NN query following the idea of ~\cite{Yang:indoorknn}.
In this paper, we use $k$NN in indoor environment to imply probabilistic $k$NN. 
\begin{definition}
\label{def:kNN} (\textit{Probabilistic k Nearest Neighbor Queries}) 
Given a set of indoor moving objects $O$=$\{o_1,o_2,..,o_n\}$,
a $k$NN query issued at time $t$ with query location
$q$ returns a result set \emph{R} = $\{A | A  \subseteq O \wedge prob(A, q, t) > k \wedge |A| \leq |A'| (\forall A' \subseteq O \wedge prob(A', q, t)>k ) \}$. We denote the probability that object $a$ exists at the searched location at time $t$ by $prob(a, t)$ (while the searching depends on the relative distance to $q$), and the total probability of all objects in the result set by $prob(A, q, t)=\sum_{a \in A} prob(a, t)$.  

\end{definition}

\begin{definition}
\label{def:range} (\textit{Range Queries}) 
Given a set of indoor moving objects $O$ = $\{o_1, o_2, ..., o_n\}$ and a range
$r$, a range query issued at time $t$ with query location
$q$ returns a result set \emph{R} = $\{A | A  \subseteq O \wedge A \in r \}$,
and the respective probabilities of each $o \in A$. 
\end{definition}

\section{System Design}\label{sec:design}
In this section, we will introduce the design of an RFID-based
indoor range and $k$NN query evaluation system, which incorporates
four modules: event-driven raw data collector, query aware
optimization module, Bayesian filtering-based preprocessing
module, and query evaluation module. In addition, we introduce the
underlying framework of two models: \emph{indoor walking graph
model} and \emph{anchor point indexing model}. We will elaborate
on the function of each module and model in the following
subsections.

\begin{figure}[t!]
\vspace{-2mm}
  \centering 
  \includegraphics[width=0.7\textwidth]{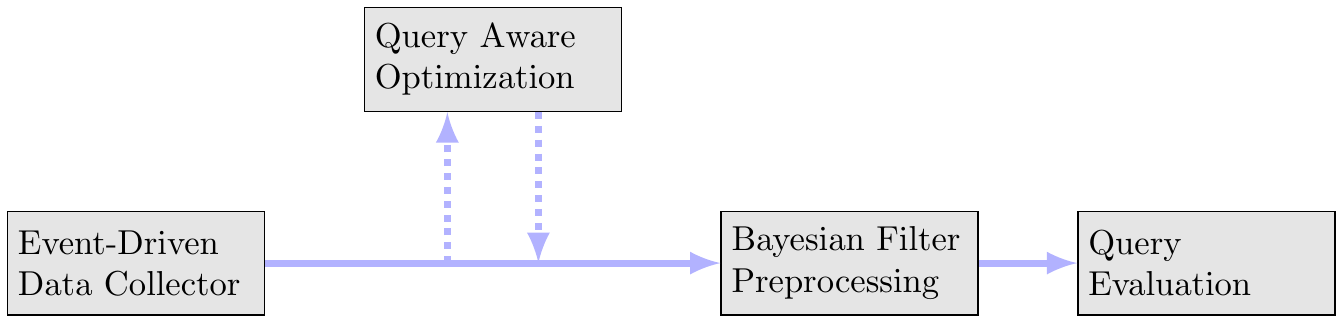}
  \caption{Overall system structure.}
  \label{figure:overall}
  \vspace{-2mm}
\end{figure}

Figure~\ref{figure:overall} shows the overall structure of our
system design. Raw readings are first fed into and processed by
the event-driven raw data collector module, which then provides
aggregated readings for each object at every second to the
Bayesian filtering-based preprocessing module. Before running the
preprocessing module, the reading data may be optionally sent to
the query aware optimization module which filters out
non-candidate objects according to registered queries and objects'
most recent readings, and outputs a candidate set $C$ to the
Bayesian filtering-based preprocessing module. The preprocessing
module cleanses noisy raw data for each object in $C$, stores the
resulting probabilistic data in a hash table, and passes the hash
table to the query evaluation module. At last, the query
evaluation module answers registered queries based on the hash
table that contains filtered data.

\subsection{Event-Driven Raw Data Collector}

In this subsection, we describe the event-driven raw data
collector which is the front end of the entire system. The data
collector module is responsible for storing RFID raw readings in
an efficient way for the following query processing tasks.
Considering the characteristics of Bayesian filtering, readings of
one detecting device alone cannot effectively infer an object's
moving direction and speed, while readings of two or more
detecting devices can. We define events in this context as the
object either entering (ENTER event) or leaving (LEAVE event) the
reading range of an RFID reader. To minimize the storage space for
every object, the data collector module only stores readings
during the most recent {ENTER, LEAVE, ENTER} events, and removes
earlier readings. In other words, our system only stores readings
of up to the two most recent consecutive detecting devices for
every object. For example, if an object is previously identified
by $d_i$ and $d_j$ (readers), readings from $d_i$ and $d_j$ are stored in
the data collector. When the object is entering the detection
range of a new device $d_k$, the data collector will record
readings from $d_k$ while removing older readings from $d_i$. The
previous readings have negligible effects on the current
prediction.

The data collector module is also responsible for aggregating the
raw readings to more concise entries with a time unit of one
second. RFID readers usually have a high reading rate of tens of
samples per second. However, Bayesian filtering does not need such
a high observation frequency. An update frequency of once per
second would provide a good enough resolution. Therefore,
aggregation of the raw readings can further
save storage without compromising accuracy. 

\subsection{Indoor Walking Graph Model and Anchor Point Indexing Model}
\label{sec:anchorpoints}

This subsection introduces the underlying assumptions and backbone
models of our system, which form the basis for understanding
subsequent sections. We propose two novel models in our system,
indoor walking graph model and anchor point indexing model, for
tracking object locations in indoor environments.

\subsubsection{Indoor Walking Graph Model}

We assume our system setting is a typical office building where
the width of hallways can be fully covered by the detection range
of sensing devices (which is usually true since the detection
range of RFID readers can be as long as 3 meters), and RFID
readers are deployed only along the hallways. In this case the
hallways can simply be modeled as lines, since from RFID reading
results alone, the locations along the width of hallways cannot be
inferred. Furthermore, since no RFID readers are deployed inside
rooms, the resolution of location inferences cannot be higher than
a single room.

\begin{figure}[b!]
  \centering
  \includegraphics[width=0.5\textwidth]{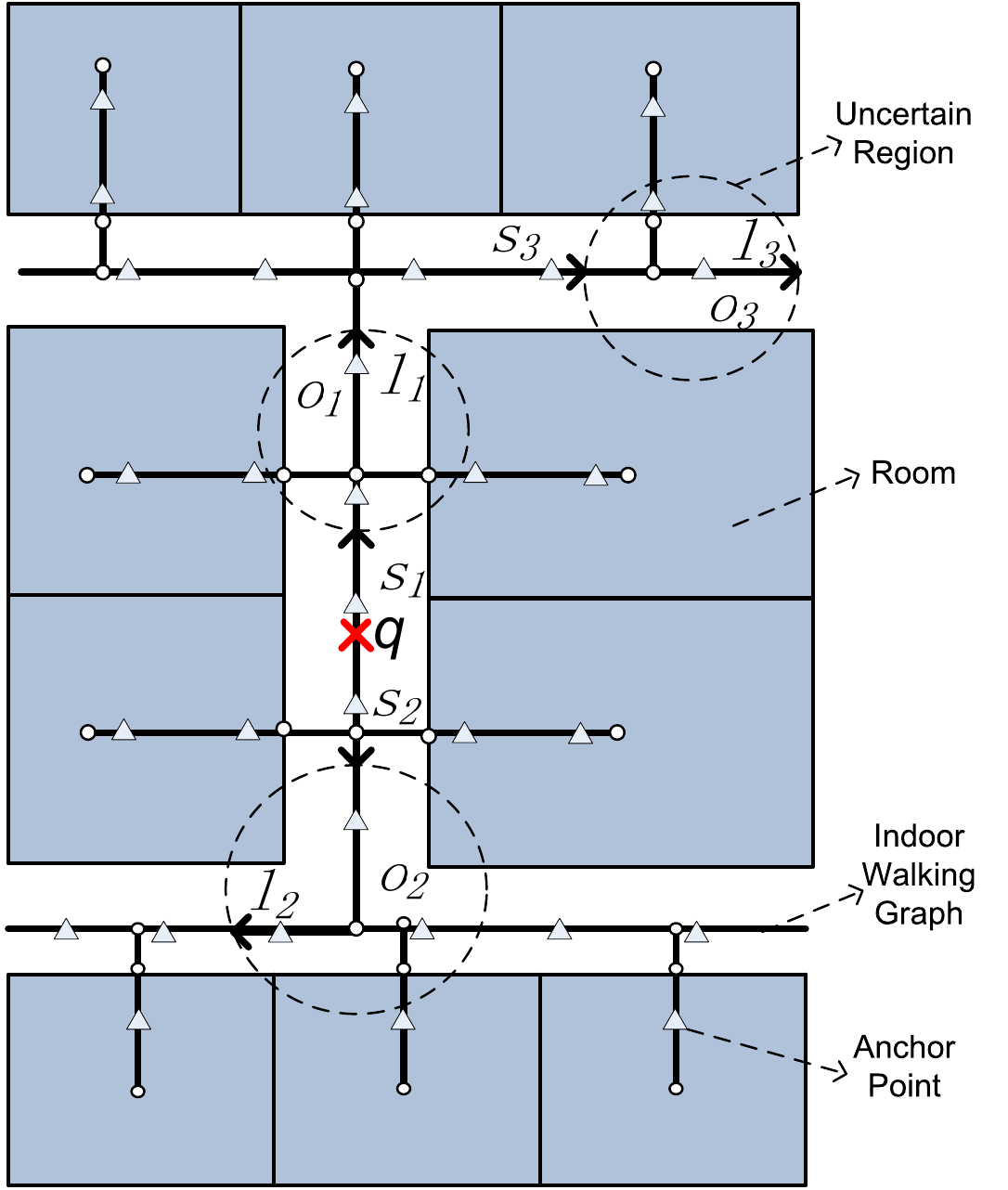}
  \caption{Filtering non-candidate objects ($k$NN query).} \label{figure:op2}
\end{figure}


Based on the above assumptions, we propose an \emph{indoor walking
graph model}. The indoor walking graph $G\langle N, E\rangle$ is
abstracted from the regular walking patterns of people in an
indoor environment, and can represent any accessible path in the
environment. The graph $G$ comprises a set $N$ of nodes (i.e.,
intersections) together with a set $E$ of edges, which present possible routes (i.e., hallways).
By restricting object movements to be only on the edges $E$ of
$G$, we can greatly simplify the object movement model while at
the same time still preserving the inference accuracy of Bayesian
filtering. Also, the distance metric used in this paper, e.g., in
$k$NN query evaluations, can simply be the shortest spatial
network distance on $G$, which can then be calculated by many
well-known spatial network shortest path
algorithms~\cite{conf/vldb/PapadiasZMT03,conf/sigmod/SametSA08} as
shown in Figure~\ref{figure:op2}.

\subsubsection{Anchor Point Indexing Model}

The indoor walking graph edges $E$ are by nature continuous. To
simplify the representation of an object's location distribution
on $E$, we propose an effective spatial indexing method: anchor
point-based indexing. We define anchor points as a set $AP$ of
predefined points on $E$ with a uniform distance (such as 1 meter)
to each other. Those anchor points are discrete location points.
For most applications, this generalization will avoid a heavy load of unnecessary computation.
An example of anchor points is shown in Figure~\ref{figure:op2}. A triangle represents an anchor point. In Figure~\ref{figure:op1}, the striped circle represents the Uncertain Region. In essence, the model of anchor points is a scheme of trying to discretize objects' locations. After Bayesian filtering is finished for an object $o_i$, its location probability distribution is aggregated to discrete anchor points.
Specifically, for the Kalman filter, an integration of an object's
bell-shaped location distribution between two adjacent anchor
points is calculated. For particle filters, suppose $ap_j$ is an
anchor point with a nonzero number $n$ of particles,
$p_i(o_i.location=ap_j)=n/N_s$, where $p_i$ is the probability
distribution function that $o_i$ is at $ap_j$ and $N_s$ is the
total number of particles for $o_i$.

A hash table \texttt{APtoObjHT} is maintained in our system.
Given the coordinates of an anchor point $ap_j$, the table will return the list of each object and its probability at the
anchor point: ($\langle o_i, p_i(ap_j)\rangle$). For instance, an
entry of \texttt{APtoObjHT} would look like: $(8.5, 6.2),
\{\langle o_1, 0.14\rangle, \langle o_3, 0.03\rangle, \langle o_7,
0.37\rangle \}$, which means that at the anchor point with coordinate
(8.5, 6.2), there are three possible objects ($o_1$, $o_3$, and
$o_7$), with probabilities of 0.14, 0.03, and 0.37, respectively.
With the help of the above anchor point indexing model, the query
evaluation module can simply refer to the hash table
\texttt{APtoObjHT} to determine objects' location distributions.

\subsection{Query Aware Optimization Module}
\label{sec:opt}

To answer every range query or $k$NN query, a naive approach is to
calculate the probability distribution of every object's location
currently in the indoor setting. However, if query ranges cover
only a small fraction of the whole area, then there will be a
considerable percentage of objects who are guaranteed not to be in
the result set of any query. We call those objects that have no
chance to be in any result set ``non-candidate objects". The
computational cost of running Bayesian filters for non-candidate
objects should be saved. In this subsection we present two
efficient methods to filter out non-candidate objects for range
query and $k$NN query, respectively.

\begin{figure}[t!]
  \centering
  \includegraphics[width=0.7\textwidth]{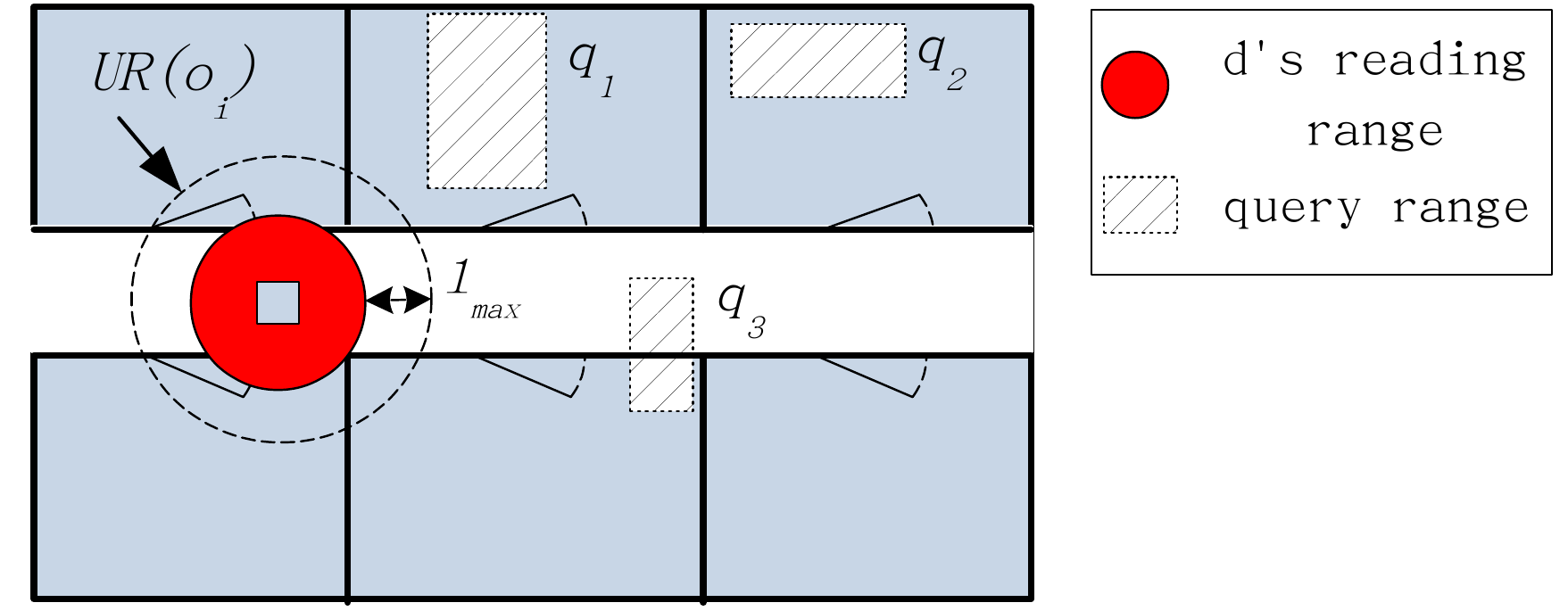}
  \caption{Filtering non-candidate objects (range query).}
  \label{figure:op1}
  \vspace{-4mm}
\end{figure}

\noindent\emph{\textbf{Range Query:}} to decrease the
computational cost, we employ a simple approach based on the
Euclidean distance instead of the minimum indoor walking
distance~\cite{Yang:indoorknn} to filter out non-candidate
objects. An example of the optimization process is shown in
Figure~\ref{figure:op1}. For every object $o_i$, its most recent
detecting device $d$ and last reading time stamp $t_{last}$ are
first retrieved from the data collector module. We assume the
maximum walking speed of people to be $u_{max}$. Within the time
period from $t_{last}$ to the present time $t_{current}$, the
maximum walking distance of a person is
$l_{max}=u_{max}*(t_{current}-t_{last})$. We define $o_i$'s
uncertain region $UR(o_i)$ to be a circle centered at $d$ with
radius $r=l_{max}+d.range$. The red circle in Figure~\ref{figure:op1} represents the reading range of a reader. If $UR(o_i)$ does not overlap with any
query range then $o_i$ is not a candidate and should be filtered
out. On the contrary, if $UR(o_i)$ overlaps with one or more query
ranges then we add $o_i$ to the result candidate set $C$. In
Figure~\ref{figure:op1}, the only object in the figure should be
filtered out since its uncertain region does not intersect with
any range query currently evaluated in the system.

\noindent\emph{\textbf{$k$NN Query:}} by employing the idea of
distance-based pruning in~\cite{Yang:indoorknn}, we perform a
similar distance pruning for $k$NN queries to identify candidate
objects. We use $s_i (l_i)$ to denote the minimum (maximum)
shortest network distance (with respect to the indoor walking
graph) from a given query point $q$ to the uncertain region of
$o_i$:

\begin{equation}
  \begin{split}
   \small
    s_i&=\min_{p\in UR(o_i)} d_{shortestpath}(q, p)\\
    l_i&=\max_{p\in UR(o_i)} d_{shortestpath}(q, p)
  \end{split}
\end{equation}

Let $f$ be the $k$-th minimum of all objects' $l_i$ values. If
$s_i$ of object $o_i$ is greater than $f$, object $o_i$ can be
safely pruned since there exists at least $k$ objects whose entire
uncertain regions are definitely closer to $q$ than $o_i$'s
shortest possible distance to $q$. Figure~\ref{figure:op2} is an
example pruning process for a 2NN query: There are 3 objects in
total in the system. We can see $l_1<l_2<l_3$ and consequently
$f=l_2$ in this case; $s_3$ is greater than $f$, so $o_3$ has no
chance to be in the result set of the 2NN query. We run the
distance pruning for every $k$NN query and add possible candidate
objects to $C$.

Finally, a candidate set $C$ is produced by this module,
containing objects that might be in the result set of one or more
range queries or $k$NN queries. $C$ is then fed into the Bayesian
filtering-based preprocessing module which will be explained in
the next subsection.

\subsection{Bayesian Filtering-based Preprocessing Module}

The preprocessing module estimates an object's location
distribution according to its two most recent readings, calculates
the discrete probability on anchor points, and stores the results
to the hash table \texttt{APtoObjHT}. We introduce two
preprocessing approaches based on two famous algorithms in the
Bayesian Filtering family: the Kalman filter and the Particle
filter.

\subsubsection{Kalman Filter-Based Preprocessing Module}

\begin{figure}[t!]
  \centering
  \vspace{-2mm}
  \includegraphics[width=0.7\textwidth]{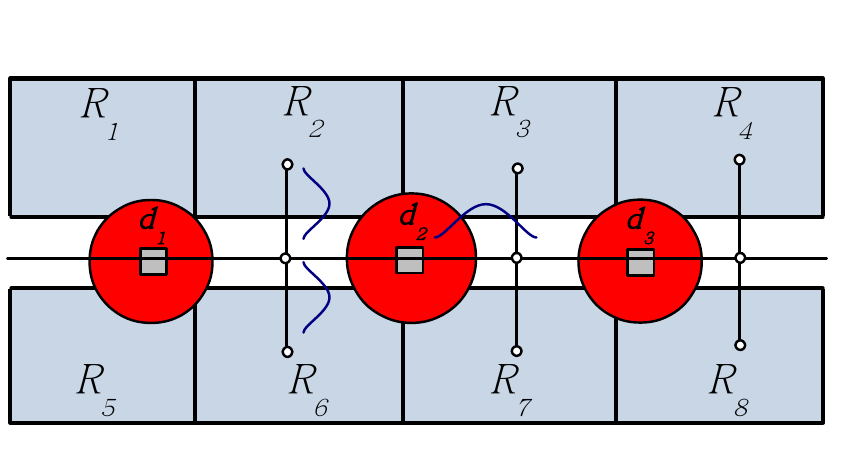}
  \vspace{-2mm}
  \caption{Kalman filter-based prediction.}
  \label{figure:kf}
  \vspace{-4mm}
\end{figure}

In this section, we extend the basic 1-D example of the Kalman
filter in Section~\ref{sec:KFprelim} to be suitable for more
complex 2-D indoor settings. Due to the irregularity of indoor
layout, the main challenge here is that an object's moving path
may diverge to multiple paths. For example, in
Figure~\ref{figure:kf}, assume an object was detected first by
reader $d_1$ at $t_1$ then by reader $d_2$ at $t_2$, it could have
entered $R_2$ or $R_6$ before proceeding to $d_2$. When we conduct
a prediction with the Kalman filter, we need to consider all these
possible paths, each of which will give a separate prediction.
Algorithm~\ref{alg:Kalman} formulates our approach of applying the
Kalman filter to estimate objects' locations, which is elucidated
in the rest of this subsection with the example in
Figure~\ref{figure:kf}.

The Kalman filter algorithm starts by first retrieving the most recent
readings for each candidate from the data collector module. Line 5
of Algorithm~\ref{alg:Kalman} restricts the Kalman filter from
running more than 60 seconds beyond the last active reading, since
otherwise its location estimation will become dispersed over a
large area and the filtering result will become unusable.

We assume objects' speed $v$ is a Gaussian variable with $\mu=1$
m/s and $\sigma=0.1$ m/s, and the time of an object staying inside a
room $t_{room}$ also follows Gaussian distribution. We assume that objects rarely enter the same room more than
once. There could be several shortest paths from reader $d_1$ to $d_2$. For a specific shortest path, if object can walk into 0 rooms, 1 room, 2 rooms, 3 rooms... m rooms during $t_1$ to $t_2$, there are $m+1$ different predictions
$\hat{x}_{2^-}=\hat{x}_1+v*(t_2-t_1-i*\mu_{t_{room}})$. We calculate the possibilities respectively on these cases from line 6 to line 16. Note that we simplify $\hat{x}_{2^-}$ by
replacing $t_{room}$ with its mean value $\mu_{t_{room}}$. For example, in Figure ~\ref{figure:kf}, the object could enter 0 rooms, 1 room, 2 rooms while moving before entering  $d_2$'s range, therefore, there are 3 distributions (0 rooms, 1 room, 2 rooms). The 3 curves in Figure ~\ref{figure:kf} indicate 3 distributions.
\begin{algorithm}[!t]
    \caption{Kalman Filter($C$)}
    \label{alg:Kalman}
    \small
\begin{algorithmic}[1]
    \FOR {each object $o_i$ of $C$}
      \STATE retrieve $o_i$'s aggregated readings from the data collector module
      \STATE $t_1$, $t_2$ = the starting/ending time of the aggregated readings
      \STATE $d_1$, $d_2$ = the second most/most recent detecting devices for $o_i$
      \STATE $t_{min}$ = min($t_2+60, t_{current}$)
      \STATE $p$ = number of shortest paths from $d_1$ to $d_2$
      \STATE $\widehat p$ = $p * m$
      \FOR {$i=1,\ldots,p$}
          \STATE $m$ = number of rooms on path $i$ from $d_1$ to $d_2$
          \STATE $\hat{x}_1$ = the mean of $o_i$'s position distribution at time $t_1$
          \STATE $\mu_{t_{room}}$ = mean time of the object staying in the room.
          \FOR {$j=0,\ldots,m$}
              \STATE $\hat{x}_{2^-}=\hat{x}_1+v*(t_2-t_1-i*\mu_{t_{room}})$
              \STATE $\sigma_{2^-}^2=\sigma_1^2+\sigma_v^2*(t_2-t_1)$
              \IF {this distribution's overlap with $d_2$'s range is below threshold}
                \STATE $\widehat p$ = $\widehat p$ - 1
              \ELSE
              \STATE Add the current route to the set of possible routes
              \ENDIF
          \ENDFOR
    \ENDFOR
    
\STATE assign the probability of each valid route with $1/\widehat p$
      \STATE calculate $\hat{x}_2$ and $\sigma_2^2$ by employing Equations~\ref{eq:kalmanmean} and~\ref{eq:kalmanvar}
      \STATE recursively enumerate all possible routes from $\hat{x}_2$ going forward until $t_{min}$
      \STATE estimate $o_i$'s location $\hat{x}_{min^-}$ by counting
      \STATE $\sigma_{min^-}^2=\sigma_2^2+\sigma_v^2*(t_{min}-t_2)$
      \FOR {each anchor point $ap_j$ with a non-zero probability under estimated location distribution}
        \STATE assign $ap_k$ as the adjacent point of $ap_j$ on the moving direction ($d_1$ to $d_2$)
        \STATE $p_i(o_i.location=ap_j)$=$\int_{ap_j}^{ap_k}p_i({x}_{min^-})$
        \STATE update Hash Table \texttt{APtoObjHT}
      \ENDFOR
    \ENDFOR
    \RETURN possible objects and their probabilities for each anchor
\end{algorithmic}
\end{algorithm}
When the observation at $t_2$ is made, we combine the observation
with only reasonable predictions to get a final estimation. By
``reasonable", we mean predictions with a good portion of pdf
overlapping with $d_2$'s reading range. For example, in
Figure~\ref{figure:kf}, if the threshold $\tau$ about the probability of the object being in $d_2$'s range is 0.05 and the probability that the object moving into $R_2$ and $R_6$ before being in $d_2$'s range is less than 0.05, this path will be eliminated. It means two predictions for the two paths entering $R_2$ and $R_6$ respectively are hardly overlapping with
$d_2$'s reading range, so we can safely prune them and only
consider the rightmost prediction. After pruning, the average of
remaining predictions is used to calculate the object's location
estimation at $t_2$ according to Equations~\ref{eq:kalmanmean}
and~\ref{eq:kalmanvar}. For example, if the distance from $d_1$ to $d_2$ is 10, the observed mean $z_2$ will be 10 and the variance is 2 (the radius of the reader’s detection range). Suppose that the predicted mean  ${\hat{x}}_{2-}$ is 14 and variance $\sigma_{2^-}^2$ is 3. By employing Equation $K_1=\sigma_{1^-}^2/(\sigma_{1^-}^2+\sigma_{z_1}^2)$, $K_1$ will be 0.6. According to Equations~\ref{eq:kalmanmean}
and~\ref{eq:kalmanvar}, the filtered mean is 11.6 and the new variance is 1.2. 

From the latest detected time $t_2$ to current, the object can
take every possible path from $d_2$ going forward. Line 19 uses
recursion to enumerate all the possibilities and line 20
calculates the probability distribution of $\hat{x}_{min^-}$. Suppose that $t_2$ is 20 and $t_{min}$ is 22.5; $t_{min}-\ t_2$ will be 2.5. In line 21, we could arrive at the new variance 1.45. From line 22 to line 26, we calculate possible objects and the integration of their probabilities. Suppose that we are going to calculate object $O_1$'s probability for anchor point ${ap}_j$ to ${ap}_k$, where ${ap}_k$ is the adjacent point in the moving direction on a specific path. If the distance from ${ap}_j$ to reader $d_1$ is 12, and the distance from ${ap}_k$ to reader $d_1$ is 16. The integration of probability is $cdf(16) - cdf(12)$, where $cdf()$ is the cumulative distribution function for the distribution calculated from lines 18 to 21. In line 25, we update the hash table APtoObjHT for each anchor. For example, there are 3 possible objects for anchor point ${ap}_{12}$: $O_1$ with probability $0.5$, $O_3$ with probability $0.9$, and $O_{15}$ with probability $0.8$. We will update the hash table with item $({ap}_{12}, {<O_1, 0.5>, <O_3, 0.9>, <O_{15}, 0.8>})$. With the aforementioned approach, we could determine possible objects and their probabilities for each anchor.

\subsubsection{Particle Filter-Based Preprocessing Module}\label{sec:preprocessing}


The particle filter method consists of 3 steps: initialization,
particle updating, and particle resampling. In the first step, a
set of particles is generated and uniformly distributed on the
graph edges within the detection range of $d_2$, and each particle
picks its own moving direction and speed as in line 5. In our
system, particles' speeds are drawn from a Gaussian distribution
with $\mu=1$ m/s and $\sigma=0.1$ m/s . In the location updating step
in line 9, particles move along graph edges according to their
speed and direction, and will pick a random direction at
intersections; if particles are inside rooms, they continue to
stay inside with probability 0.9 and move out with probability
0.1. After location updating, in line 16, weights of particles are
updated according to their consistency with reading results. In
other words, particles within the detecting device's range are
assigned a high weight, while others are assigned a low weight. In
the resampling step, particles' weights are first normalized as in
line 18. We then employ the Resampling
Algorithm~\cite{conf/edbt/YuKSL13} to replicate highly weighted
particles and remove lowly weighted particles as in line 19. Lines
23 to 26 discretize the filtered probabilistic data and build the
hash table \texttt{APtoObjHT} as described in Section~\ref{sec:anchorpoints}.
\begin{algorithm}[!t]
    \algsetup{linenosize=\small,linenodelimiter=.}
    \caption{Particle Filter($C$)}
    \label{alg:PF}
    \small
\begin{algorithmic}[1]
    \FOR {each object $o_i$ of $C$}
      \STATE retrieve $o_i$'s aggregated readings from the data collector module
      \STATE $t_1$, $t_2$ = the starting/ending time of the aggregated readings
      \STATE $d_1$, $d_2$ = the second most/most recent detecting devices for $o_i$
      \STATE initialize particles with random speed and direction within $d_2$'s activation range
      \STATE $t_{min}$ = min($t_2+60, t_{current}$)
      \FOR {every second $t_j$ from $t_1$ to $t_{min}$}
          \FOR {every particle $p_m$ of $o_i$}
         \STATE $p_m$ updates its location
         \ENDFOR
         \STATE retrieve the aggregated reading entry \emph{reading} of $t_j$
         \IF {$reading.Device$=\emph{null}}
            \STATE continue
         \ELSE
            \FOR {every particle $p_m$ of $o_i$}
               \STATE update $p_m$'s weight
            \ENDFOR
            \STATE normalize the weights of all particles of $o_i$
            \STATE Resampling() 
         \ENDIF
      \ENDFOR
      \STATE assign particles of $o_i$ to their nearest anchor points
      \FOR {each anchor point $ap_j$ with a nonzero number of particles $n$}
        \STATE calculate probability $p_i(o_i.location=ap_j)=n/N_s$
        \STATE update Hash Table \texttt{APtoObjHT}
      \ENDFOR
    \ENDFOR
    \RETURN possible objects and their possibilities
\end{algorithmic}
\end{algorithm}
\subsection{Query Evaluation}\label{sec:queryEva}

In this subsection we are going to discuss how to evaluate range
and $k$NN queries efficiently with the filtered probabilistic data
in the hash table \texttt{APtoObjHT}. For $k$NN queries, without
loss of generality, the query point is approximated to the nearest
edge of the indoor walking graph for simplicity.

\subsubsection{Indoor Range Query}

\begin{figure}[b]
  \centering
  \includegraphics[width=0.5\textwidth]{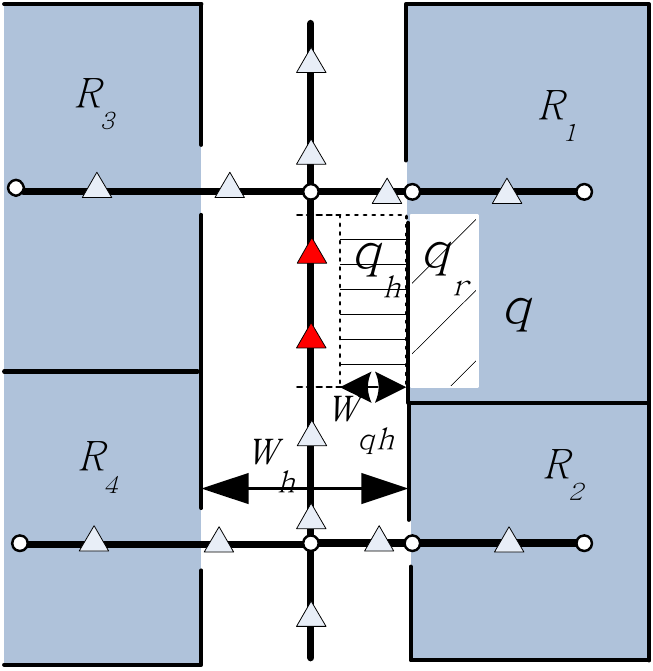}
  \caption{Example of indoor range query.}
  \label{figure:evaluation}
\end{figure}

To evaluate indoor range queries, the first thought would be to
determine the anchor points within the range, then answer the
query by returning objects and their associated probabilities
indexed by those anchor points. However, with further
consideration, we can see that since anchor points are restricted
to be only on graph edges, they are actually the 1-D projection of
2-D spaces; the loss of one dimension should be compensated in the
query evaluation process. Figure~\ref{figure:evaluation} shows an
example of how the compensation is done with respect to two
different types of indoor entities: hallways and rooms.

\begin{algorithm}[!t]
    \algsetup{linenosize=\small,linenodelimiter=.}
    \caption{Indoor Range Query($q$)}
    \label{alg:range}
    \small
\begin{algorithmic}[1]
\STATE $resultSet$=$\emptyset$ \STATE $cells$=getIntersect($q$) \FOR
{every $cell$ in $cells$}
    \IF{$cell.type$=HALLWAY}
    \STATE $anchorpoints=cell.$getCoveredAP($q$)
    \STATE $ratio=cell$.getWidthRatio($q$)
    \ELSIF{$cell.type$=ROOM}
    \STATE $anchorpoints=cell$.getInsideAP()
    \STATE $ratio=cell.$getAreaRatio($q$)
    \ENDIF
    \STATE $result$=$\emptyset$
    \FOR{each $ap$ in $anchorpoints$}
       \STATE $result$=$result$+APtoObjHT.get($ap$)
    \ENDFOR
    \STATE $result=result*ratio$
    \STATE $resultSet=resultSet+result$ 
\ENDFOR \RETURN $resultSet$  \COMMENT{ All the objects in range and their possibilities }
\end{algorithmic}
\end{algorithm}

In Figure~\ref{figure:evaluation}, query $q$ is a rectangle which
intersects with both the hallway and room $R_1$, but does not
directly contain any anchor point. We denote the left part of $q$
which overlaps with the hallway as $q_h$, and the right part which
overlaps with $R_1$ as $q_r$. We first look at how to evaluate the
hallway part of $q$. The anchor points which fall within $q$'s
vertical range are marked red in Figure~\ref{figure:evaluation},
and should be considered for answering $q_h$. Since we assume there is no differentiation along the width of hallways, objects in hallways can
be anywhere along the width of hallways with equal probability.
With this assumption, the ratio of $w_{q_h}$ (the width of $q_h$)
and $w_h$ (the width of the hallway) will indicate the probability
of objects in hallways within the vertical range of $q$ being in
$q_h$. For example, if an object $o_i$ is in the hallway and in
the vertical range of $q$ with probability $p_1$, which can be
calculated by summing up the probabilities indexed by the red
anchor points, then the probability of this object being in $q_h$
is $p_i(o_i.location\in q_h)=p_1*w_{q_h}/w_h$.

Then we look at the room part of $q$. The anchor points within
room $R_1$ should represent the whole 2-D area of $R_1$, and again
we assume objects inside rooms are uniformly distributed. Similar
to the hallway situation, the ratio of $q_r$'s area to $R_1$'s
area is the probability of an object in $R_1$ happening to be in
$q_r$. For example, if $o_i$'s probability of being in $R_1$ is
$p_2$, then its probability of being in $q_r$ is
$p_i(o_i.location\in q_r)=p_2*Area_{q_r}/Area_{R_1}$, where $p_2$
can be calculated by summing up the indexed probabilities of $o_i$
on all the anchor points inside $R_1$, and $Area_i$ stands for the
size of a given region $i$.

Algorithm~\ref{alg:range} summarizes the above procedures. In line
15, we define the multiplication operation for $resultSet$ to adjust the
probabilities for all objects in it by the multiplying constant.
In line 16, we define the addition operation for $resultSet$:
if an object probability pair $\langle o_i, p\rangle$ is to be
added, we check whether $o_i$ already exists in $resultSet$. If so,
we just add $p$ to the probability of $o_i$ in $resultSet$;
otherwise, we insert $\langle o_i, p\rangle$ to $resultSet$. For
instance, suppose $resultSet$ originally contains $\{(o_1, 0.2),
(o_2, 0.15)\}$, and result stores $\{(o_2, 0.1), (o_3, 0.05)\}$. After the addition in line 16, $resultSet$ is updated to be $\{(o_1, 0.2), (o_2, 0.25), (o_3,
0.05)\}$.

\subsubsection{Indoor $k$NN Query}

For indoor $k$NN queries, we present an efficient evaluation
method with statistical accuracy. Unlike previous
work~\cite{Yang:indoorknn,Cheng:2009:EPT:1516360.1516438}, which
involves heavy computation and returns multiple result sets for
users to choose, our method is user friendly and returns a
relatively small number of candidate objects. Our method works as
follows: starting from the query point $q$, anchor points are
searched in ascending order of their distance to $q$; the search
expands from $q$ one anchor point forward per iteration, until the
sum of the probability of all objects indexed by the searched
anchor points is no less than $k$. The result set has the form of
$\langle(o_1, p_1), (o_2, p_2), ... (o_m, p_m)\rangle$ where
$\sum_{i=1}^{m} p_i \geq k$. The number of returned objects will
be at least $k$. From the sense of statistics, the probability
$p_i$ associated with object $o_i$ in the result set is the
probability of $o_i$ being in the $k$NN result set of $q$. The
algorithm of the indoor $k$NN query evaluation method in our work
is shown in Algorithm~\ref{alg:kNN}.

\begin{algorithm}[!t]
    \algsetup{linenosize=\small,linenodelimiter=.}
    \caption{Indoor $k$NN Query($q$, $k$)}
    \label{alg:kNN}
    \small
\begin{algorithmic}[1]
\STATE $resultSet$=$\emptyset$

\STATE $\overline{n_in_j}$=find\_segment($q$)

\STATE vector $V$=$\langle(n_i,q), (n_j,q)\rangle$  // elements in $V$
have the form ($node$, $prevNode$) \FOR {every entry $e$ in $V$}
  \STATE $anchorpoint$=find\_nextAnchorPoint($e$) // return the next unsearched anchor point from $e.prevNode$ to $e.node$
  \IF{$anchorpoint$=$\emptyset$}
    \STATE remove $e$ from $V$
    \FOR{each unvisited adjacent node $n_x$ of $e$.node}
       \STATE add ($n_x$, $e.node$) to $V$
    \ENDFOR
    \STATE continue
  \ENDIF
  \STATE $resultSet$=$resultSet$+APtoObjHT.get($anchorpoint$)
  \STATE $prob_{total}$=$resultSet$.getTotalProb() 
  \IF{$prob_{total} >= k$}
    \STATE break
  \ENDIF
\ENDFOR \RETURN $resultSet$
\end{algorithmic}
\end{algorithm}

In Algorithm~\ref{alg:kNN}, lines 1 and 2 are initial setups. Line
3 adds two entries to a vector $V$, whose elements store the edge
segments expanding out from query point $q$. In the following \textbf{for}
loop, line 5 finds the next unvisited anchor point further away
from $q$. If all anchor points are already searched on an edge
segment $e$, lines 6 to 12 remove $e$ and add all adjacent
unvisited edges of $e$.node to $V$. Line 13 updates the result set
by adding $\langle$object ID, probability$\rangle$ pairs indexed
by the current anchor point to it. In lines 14 to 17, the total
probability of all objects in the result set is checked, and if it
equals or exceeds $k$, the algorithm ends and returns the result
set. Note that the stopping criteria of our $k$NN algorithm do not
require emptying the frontier edges in $V$.

\begin{figure}[b]
  \centering
  \includegraphics[width=0.5\textwidth]{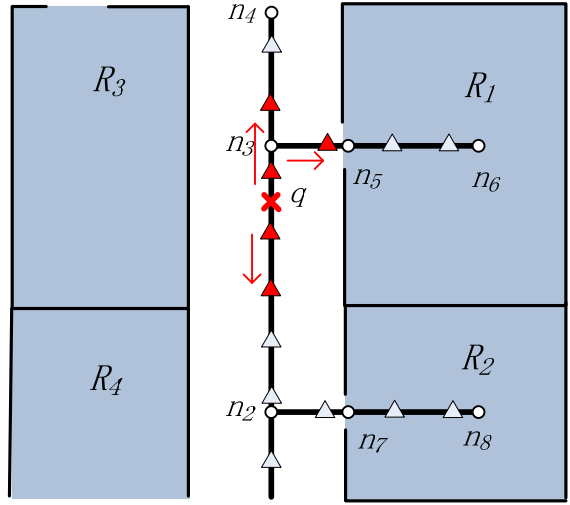}
  \caption{indoor $k$NN query.}
  \label{figure:knn}
\end{figure}

An example $k$NN query is shown in Figure~\ref{figure:knn}, which
is a snapshot of the running status of Algorithm~\ref{alg:kNN}. In
Figure~\ref{figure:knn}, red arrows indicate the searching
directions expanding from $q$, and red anchor points indicate the
points that have already been searched. Note that the edge segment
from $q$ to $n_3$ is already removed from $V$ and new edges
$\overline{n_3n_4}$, $\overline{n_3n_5}$ are currently in $V$ as
well as $\overline{n_2q}$. The search process is to be continued
until the total probability of the result set is no less than $k$.

\subsubsection{Continuous Indoor Range Query}

In this subsection, we aim to solve the problem of continuous
indoor range query on filtered probabilistic data. To efficiently
monitor the result set, we use a similar concept \emph{``critical
  device"} as in~\cite{Yang:indoorrange}, which can save
considerable computations rather than constantly repeating the
snapshot algorithm. We define \emph{critical devices} for a query
to be only the set of devices whose readings will affect the query
results. Our continuous monitoring algorithm is distinct from
Yang's work~\cite{Yang:indoorrange} in two aspects: First, we
leverage the Indoor Walking Graph to simplify the identification
process of critical devices. Second, the probability updating
process is Bayesian filter-based, which is more accurate and very
different in nature from Yang's approach.

\begin{figure}[b]
  \centering
  \includegraphics[width=0.5\textwidth]{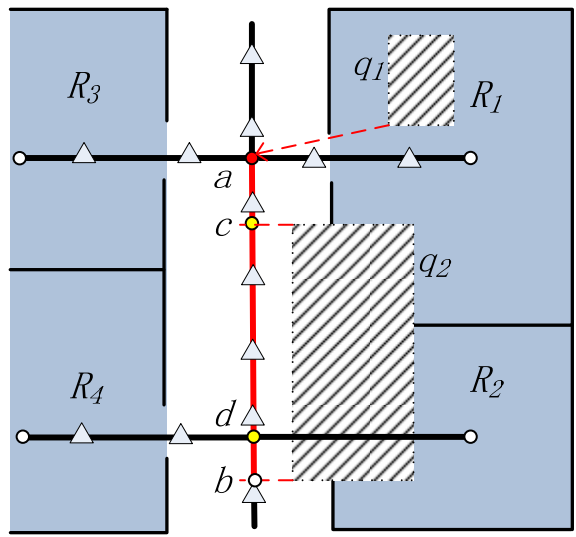}
  \caption{Mapping process to identify critical devices.}
  \label{fig:criticaldevice}
\end{figure}

To identify \emph{critical devices} for a range query, we propose
an approach consisting of two steps, mapping and searching. For
the mapping step, we categorize two different cases:

\begin{itemize}
\item\emph{Case 1}: when the whole query range is contained within one
  room or adjacent rooms, then we project from the doors of end
  rooms to $E$ along hallways. For example, $q_1$ in
  Figure~\ref{fig:criticaldevice} is fully contained in room $R_1$,
  so it is projected to a point (the red point) on $E$ through the
  door of $R_1$.

\item\emph{Case 2}: when the query range overlaps with both rooms and
  hallways, then the endpoints of mapped edge segment(s) should take
  whichever makes the covered segment longer among projected points
  of query range ends and end rooms' doors. In
  Figure~\ref{fig:criticaldevice}, $q_2$ is an example of this case. It is
  mapped to an edge segment, $\overline{ab}$, along the hallway as
  marked in red. Point $a$, room $R_1$ door's projected point, is
  chosen instead of $c$, the query range end projected point.
  Similarly, point $b$ is chosen instead of $d$.
\end{itemize}

For the searching step, an expansion starting from the mapped
endpoint(s) is performed along $E$ until the activation range of
an RFID reader or a dead end is reached.

For the initial evaluation of a query, we change the optimization
algorithm in Section~\ref{sec:opt} of the snapshot query to fully
take advantage of critical devices. For an object to be in the
query range, it must be most recently detected by a critical
device or any device that is bounded by the critical devices.
Other than the difference in identifying the candidate object set,
other parts of the initial evaluation algorithm are the same as
its snapshot counterpart. After initial evaluation, we
continuously monitor the candidate set by performing Bayesian
filters for them at every time step.


During the lifetime of a query, the candidate set may change due
to candidates moving out or non-candidates moving into the
critical device bounded region. If a candidate object is detected
by a critical device, or the object's probability of still
residing in the bounded region falls to 0, then we assume that it
is moving out and should be removed from the candidate set. On the
other hand, if a non-candidate object enters the detection range
of a critical device, we assume it is moving into the bounded
region and should be added to the candidate set.

The proposed continuous indoor range query is formalized in
Algorithm~\ref{alg:continuousRange}. Lines 1 to 6 initialize the
critical devices and candidate set for query $q$. In line 4 we use
a new hash table $DtoObj$, which maps a device to objects whose
most recent readings are from this device. Lines 9 to 20 update
the candidate set according to the readings of critical devices,
and also objects' probabilities of presence within the bounded
region. Line 21 executes Algorithms~\ref{alg:Kalman}
or~\ref{alg:PF} to update candidate objects' location distribution
probabilities. Line 22 calculates the result set using
Algorithm~\ref{alg:range}. Note that for Algorithm~\ref{alg:range}
there is no need to recompute anchor point set since it remains
unchanged until the query is unregistered from the system.

\begin{algorithm}[!t]
    \algsetup{linenosize=\small,linenodelimiter=.}
    \caption{Continuous Range Query($q$)}
    \label{alg:continuousRange}
    \small
\begin{algorithmic}[1]
\STATE $D_{cd}$=getCriticalDevices($q$) \STATE $C=\emptyset$
\FOR{every $reader$ in or bounded by $D_{cd}$}
   \STATE $C=C\bigcup DtoObj(reader)$
\ENDFOR \STATE Bayesian Filter($C$) \STATE $R_{init}$=Indoor Range
Query($q$)

\FOR{every time step from $t_{reg}$ to $t_{unreg}$}
   \FOR{every $o_i$ detected by any reader in $D_{cd}$}
       \IF{$o_i\in C$}
           \STATE $C$.remove($o_i$)
       \ELSE
           \STATE $C$.add($o_i$)
       \ENDIF
   \ENDFOR
   \FOR{every $o_i \in C$}
       \IF{$p(o_i.location \in bounded region of D_{cd})=0$}
           \STATE $C$.remove($o_i$)
       \ENDIF
   \ENDFOR
   \STATE Bayesian Filter($C$)
   \STATE $R$=Indoor Range Query($q$)

\ENDFOR
\end{algorithmic}
\end{algorithm}

\subsubsection{Continuous Indoor $k$NN Query}

Similar to continuous indoor range query, a method for updating the
candidate set of continuous indoor $k$NN query is crucial. To
reduce the overhead of computing the candidate set at every time
step, we buffer a certain number of extra candidates, and only
recompute the candidate set according to the optimization approach
in Section~\ref{sec:opt} when the total number of candidates is
less than $k$.

Recall from Section~\ref{sec:opt}, by examining the minimum
($s_i$)/maximum ($l_i$) shortest network distance from the query
point $q$ to an object's uncertain region, the snapshot
optimization approach excludes objects with $s_i>f$. Note that the
candidate set identified by this method contains at least $k$
objects (usually more than $k$). Based on this snapshot
optimization approach, we extend it to include at least $k+y$
candidates where $y$ is a user configurable parameter. Obviously,
$y$ represents a trade-off between the size of candidate set and
the recomputing frequency. We accomplish this by calculating the
$(k+y)$-th minimum $l_i$ among all objects, and use this value as
a threshold to cut off non-candidate objects.

During continuous monitoring, we need to make sure that the
candidate set gets updated accordingly when objects move away or
towards $q$. We still use critical devices to monitor candidates,
but now the critical devices may change each time the candidate
set is recomputed. The identification process of critical devices
goes like the following: after calculating the candidate set, a
search is performed from $q$ along $E$ to cover all the uncertain
regions of candidate objects, until reaching readers (critical
devices) or a dead end. As we can see, critical devices form a
bounded region where at least $k+y$ candidate objects are surely inside it.

The proposed continuous indoor $k$NN query is formalized in
Algorithm~\ref{alg:continuousKNN}. Note that in lines 13 to 16,
when the total number of candidates falls below $k$, we need to
recompute a new candidate set of at least $k+y$ objects, and
identify new critical devices accordingly.

\begin{algorithm}[!t]
    \algsetup{linenosize=\small,linenodelimiter=.}
    \caption{Continuous $k$NN Query($q$, $k$, $y$)}
    \label{alg:continuousKNN}
    \small
\begin{algorithmic}[1]
\STATE $C$=getCandidateObjects($k+y$) \STATE
$D_{cd}$=getCriticalDevices($C$) \STATE Bayesian Filter($C$) \STATE
$R_{init}$=Indoor $k$NN Query($q$, $k$)

\FOR{every time step from $t_{reg}$ to $t_{unreg}$}
   \FOR{every $o_i$ detected by any reader in $D_{cd}$}
       \IF{$o_i\in C$}
           \STATE $C$.remove($o_i$)
       \ELSE
           \STATE $C$.add($o_i$)
       \ENDIF
   \ENDFOR
   \IF{$C.count<k$}
       \STATE $C$=getCandidateObjects($k+y$)
       \STATE $D_{cd}$=getCriticalDevices($C$)
   \ENDIF
   \STATE Bayesian Filter($C$)
   \STATE $R$=Indoor $k$NN Query($q$, $k$)

\ENDFOR
\end{algorithmic}
\end{algorithm}

\section{Experimental Validation}\label{sec:expr}
In this section, we evaluate the performance of the proposed
Bayesian filtering-based indoor spatial query evaluation system
using both synthetic and real-world data sets, and compare the results with the symbolic model-based solution~\cite{Yang:indoorknn}. The proposed algorithms are
implemented in C++.  All the experiments were conducted on an
Ubuntu Linux server equipped with an Intel Xeon 2.4GHz processor
and 16GB memory.  In our experiments, the floor plan, which
is an office setting on the second floor of the Haley Center on Auburn University
campus, includes 30 rooms and 4 hallways on a single floor, in
which all rooms are connected to one or more hallways by doors\footnote{Our code, data, and the floor plan are publicly available at \color{blue}https://github.com/DataScienceLab18/IndoorToolKit.}. A
total of 19 RFID readers are deployed on hallways with uniform distance to each other. Objects are moving continuously without stopping, waiting, or making detours.

\subsection{Evaluation Metrics}
\begin{enumerate}
\item For range queries, we proposed cover
divergence to measure the accuracy of query results from the two
modules based on their similarity with the true result. Cover
divergence is used to evaluate the difference
between two probability distributions. The
discrete form of cover divergence of $Q$ from $P$ given in
Equation~\ref{eq:K-L-d} measures the information loss when $Q$ is
used to approximate $P$. As a result, in the following
experiments, smaller cover divergence indicates better accuracy of
the results with regard to the ground truth. For instance, there are 3 objects at time $t$ in the query window: $\{o_1, o_2, o_3\}$, and the predicted result $P$ is $\{(o_1, 0.9), (o_2, 0.8), (o_3, 0.7), (o_5, 0.5)\}$. $C_d(P||Q) = log (1/0.9) +log (1/0.8) + log (1/0.7) = 0.6851$.

\item For $k$NN queries, cover divergence is no longer a suitable
metric since the result sets returned from the symbolic model
module do not contain object-specific probability information.
Instead, we count the hit rates of the results returned by
the two modules over the ground truth result set. We only consider
the maximum probability result set generated by the symbolic model
module when calculating the hit rate. Given a query point $q$, there 
will be a ground truth set $R_{Truth}$ which contains $k$ nearest objects around $q$ at time $t$. The query model will also return a predicted set $R$ = $\lbrace A \mid A \subseteq O \land prob (A, q, t) > k$ , $O$ = $\lbrace$ $o_1, o_2, ..., o_n$ $\rbrace\rbrace$. The query model sums up the probabilities of the nearest neighbor in decreasing order of distance from $q$ until $prob (A, q, t)  > k$. Hit rate is formally defined in Equation~\ref{eq:Hit-rate}. For example, if $k=3$, the ground truth set is $\{o_1, o_2, o_3 \}$, and the predicted result is
$R$ = $\{(o_1, 0.9), (o_2, 0.9), (o_4, 0.8), (o_5, 0.5)\}$, $prob(A) = 0.9+0.9+0.8+0.5=3.1$. $R_{Truth}\land R = \{o_1, o_2\}$. The hit rate is 0.667.

\end{enumerate}
\begin{equation}
  \label{eq:K-L-d}
  C_d(P||Q) = \sum_{i}P(i) \ln \frac{P(i)}{Q(i)}
\end{equation}
\begin{equation}
  \label{eq:Hit-rate}
  H(q, t) = | R_{Truth} \cap R|/|R_{Truth}|
\end{equation}
In all the following experimental result figures, we use PF, KF,
and SM to represent particle filter-based
method, Kalman filter-based method, and symbolic model-based
method, respectively. 
\subsection{Synthetic Data Set}

\begin{figure}[!ht]
  \centering
  \includegraphics[width=0.8\textwidth]{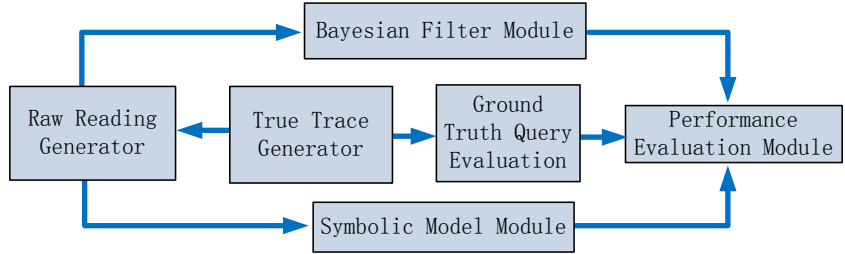}
  \caption{The simulator structure.}
  \label{figure:expr}
\end{figure}

The whole simulator consists of six components, including true
trace generator, raw reading generator, Bayesian filter module,
symbolic model module, ground truth query evaluation, and
performance evaluation module. Figure~\ref{figure:expr} shows the
relationship of different components in the simulation system. The
true trace generator module is responsible for generating the
ground truth traces of moving objects and recording the true
location of each object every second. Each object randomly selects
its destination, and walks along the shortest path on the indoor
walking graph from its current location to the destination node.
We simulate the objects' speeds using a Gaussian distribution with
$\mu=1$ m/s and $\sigma=0.1$ m/s. The raw reading generator module
checks whether each object is detected by a reader according to
the deployment of readers and the current location of the object with a certain probability.
Whenever a reading occurs, the raw reading generator will feed the
reading, including detection time, tag ID, and reader ID, to the
query evaluation modules (Bayesian filter module and symbolic
model module). The ground truth query evaluation module forms a
basis to evaluate the accuracy of the results returned by the two
aforementioned query evaluation modules. The default parameters of all the
experiments are listed in Table~\ref{tbl:DefaultValue}.

\begin{table}[!ht]
\small
\renewcommand{\arraystretch}{1.2}
\centering
\begin{tabular}{|c|c|}
\hline Parameters & Default Values \\
\hline \hline
Number of particles & 64 \\
\hline
Query window size & 2\%  \\
\hline
Number of moving objects & 200   \\
\hline
$k$ & 3   \\
\hline
Activation range & 2 meters   \\
\hline
\end{tabular}
\caption{Default values of parameters.} \vspace*{-10pt}
\label{tbl:DefaultValue}
\end{table}
\subsubsection{Effects of Query Window Size}

\begin{figure}[ht]
\vspace{-3mm}
  \begin{minipage}[t]{.45\linewidth}
   \centering
    \includegraphics[width=\textwidth]{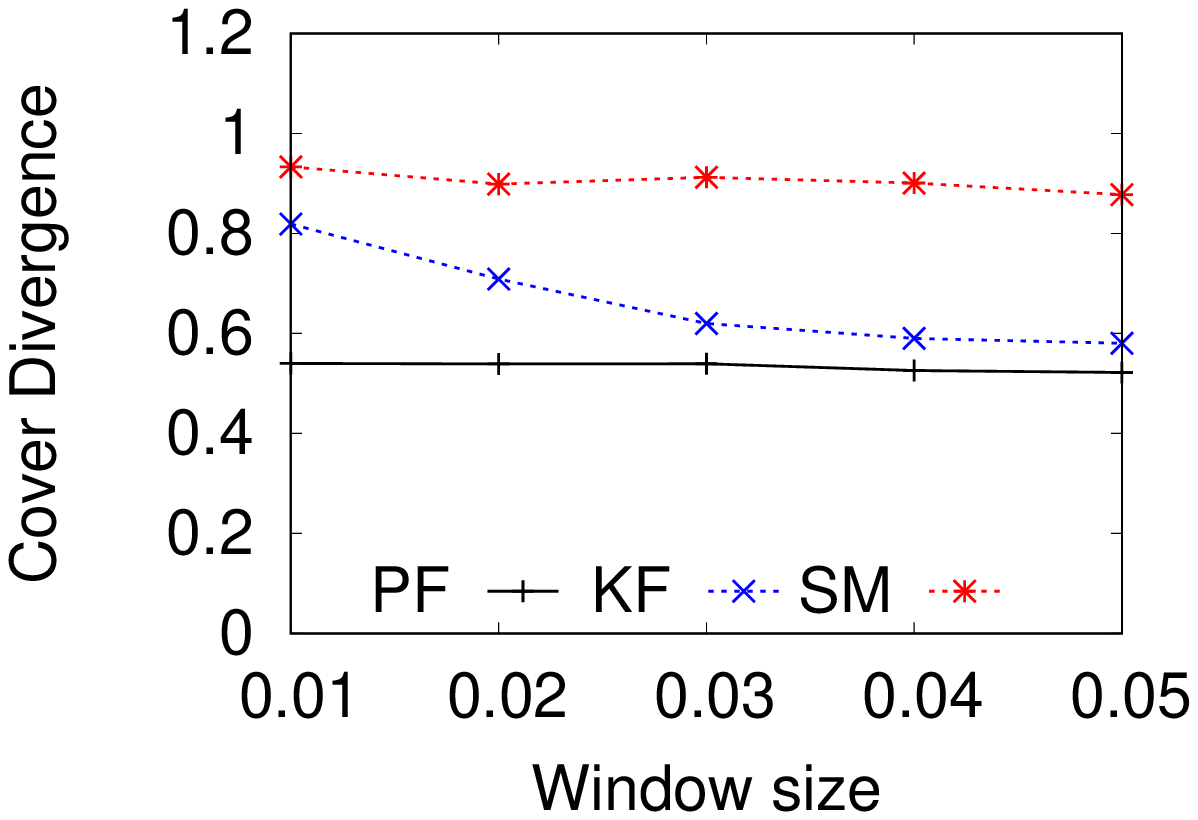}
    \caption{Effects of query window size.}
    \label{fig:queryW}
  \end{minipage}  
  \begin{minipage}[t]{.45\linewidth}
    \centering
    \includegraphics[width=\textwidth]{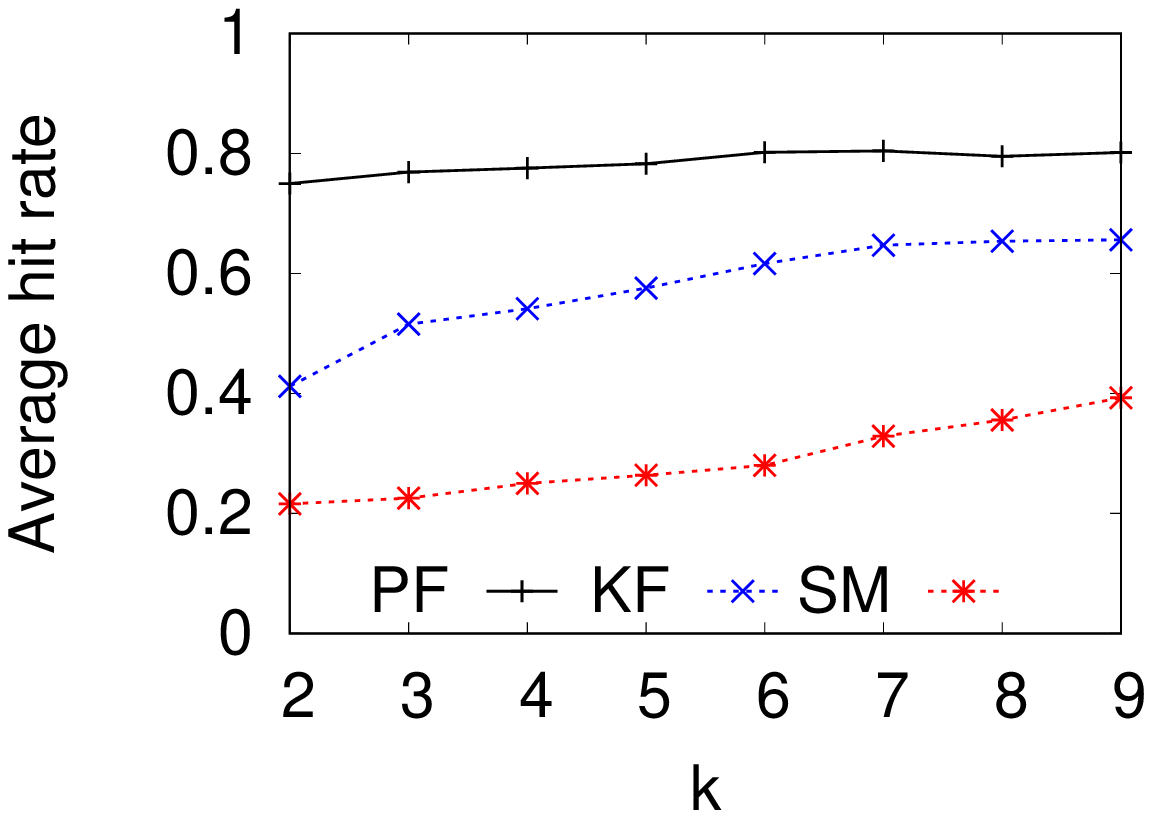}
    \caption{Effects of $k$.}
    \label{fig:knnk}
  \end{minipage}
  \vspace{-3mm}
\end{figure}

We first evaluate the effects of query window size on the accuracy
of range queries.  The window size is measured by percentage with
respect to the total area of the simulation space. At each time stamp, 100 query windows are randomly generated as rectangles, and the results are averaged over 100 different time stamps. As
shown in Figure~\ref{fig:queryW}, their accuracy is not
significantly affected by the query window size. However, the cover
divergence of the particle filter-based method is lower than both the Kalman filter-based and symbolic model-based methods.

\subsubsection{Effects of \emph{k}}

In this experiment we evaluate the accuracy of $k$NN query results
with respect to the value of $k$.  We choose 100 random indoor
locations as $k$NN query points and issue queries on these query
points at 100 different time stamps. As $k$ goes from 2 to 9, we can see in Figure~\ref{fig:knnk} that the average hit rates of Kalman filter-based and symbolic model-based methods grow slowly. As $k$ increases, the number of objects returned by the method increase as well, resulting in a higher chance of hits. On the contrary, the average hit rate of the particle filter-based method is relatively stable with respect to the value of $k$, and the particle filter-based method always outperforms the other two methods in terms of the average hit rate.

\begin{figure}
  \centering
  \vspace*{-5pt}
  \begin{subfigure}{.5\linewidth}
    \centering
    \includegraphics[width=\textwidth]{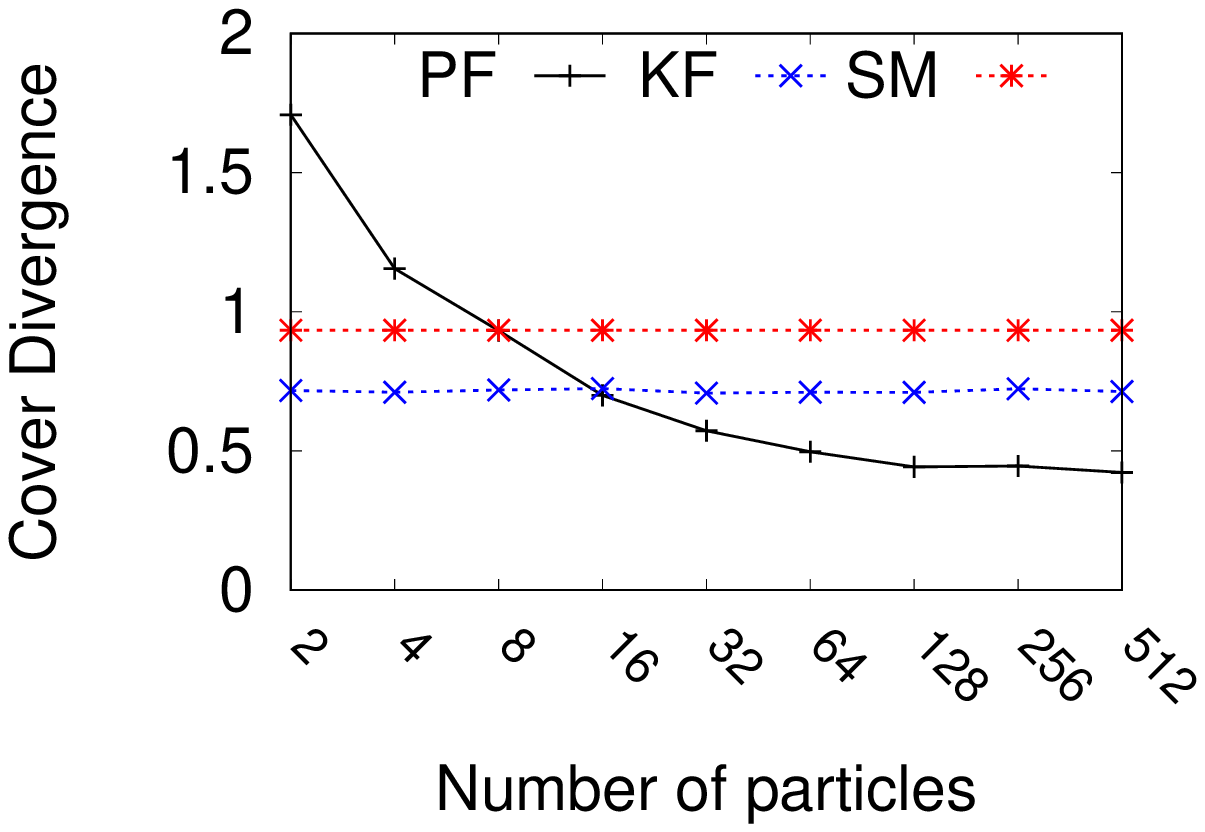}
    \caption{Cover divergence}
  \end{subfigure}%
  \begin{subfigure}{.5\linewidth}
    \centering
    \includegraphics[width=\textwidth]{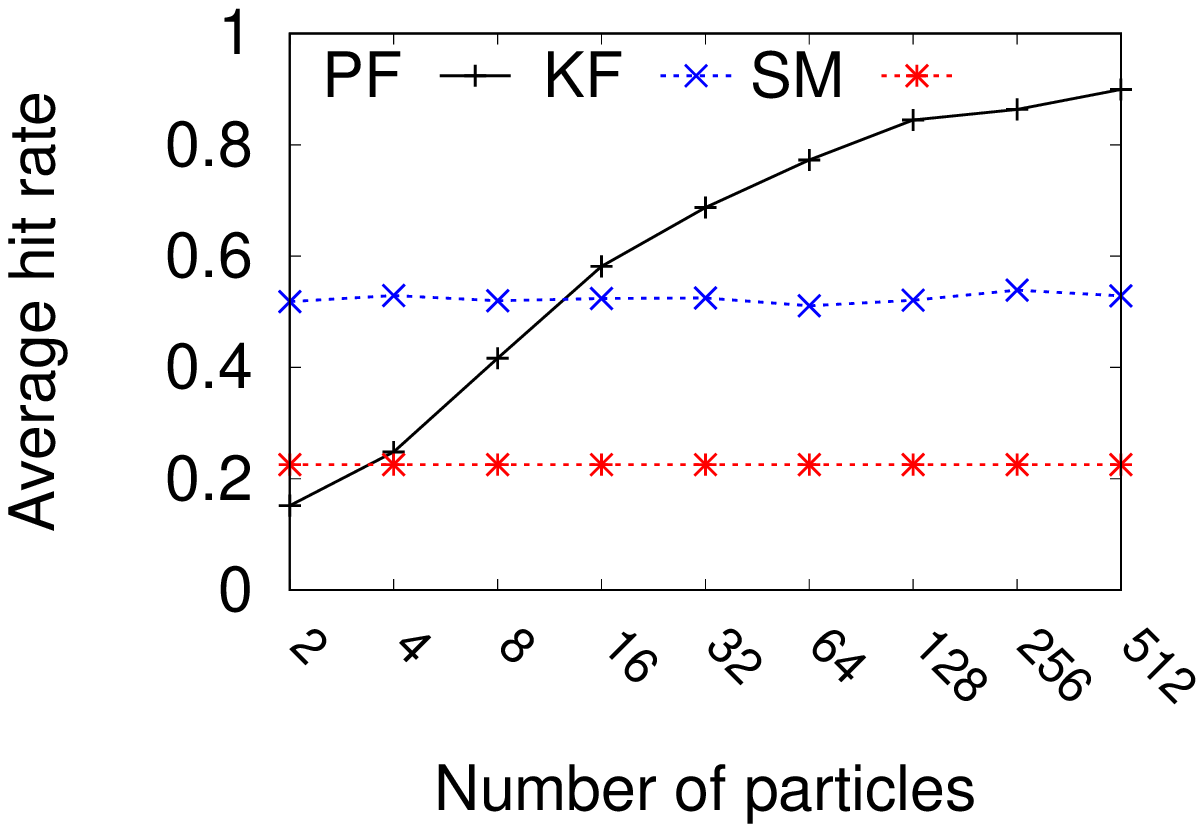}
    \caption{$k$NN success ratio}
  \end{subfigure}
  \caption{The impact of the number of particles.}
  \label{fig:numP}
\end{figure}
\begin{figure}
  \centering
   \vspace{-3mm}
  \includegraphics[width=2.5in]{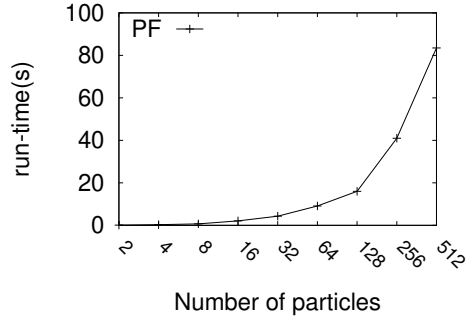}
  \caption{Run-time for different number of particles.}
  \label{fig:runtime}
   \vspace{-5mm}
\end{figure}

\subsubsection{Effects of Number of Particles}

From the mathematical analysis of particle filters in
Section~\ref{sec:pf}, we know that if the number of particles is
too small, the accuracy of particle filters will degenerate due to
insufficient samples.  On the other hand, keeping a large number of
particles is not a good choice either since the computation cost
may become overwhelming, as the accuracy improvement is no longer
obvious when the number of particles is beyond a certain
threshold.  In this subsection, we conduct extensive experiments
to explore the effects of the number of particles on query result
accuracy in order to determine an appropriate size of the particle
set for the application of indoor spatial queries.

As shown in Figure~\ref{fig:numP}, we can see that when the number of particles is very small, the particle filter-based method has a smaller average hit rate for $k$NN queries than the other two methods.  As the number of particles grows beyond 16, the performance of the particle filter-based method exceeds the other two. For range queries, the particle filter-based method has a lower cover divergence than the other two methods when the number of particles grows beyond 16. However, the performance gain with more than 64 particles slows down as we already have around $90\%$ accuracy. Figure~\ref{fig:runtime} shows the relationship between run-time and the number of particles. As the number of particles increases, the run-time increases. Therefore, we conclude that in our application, the appropriate size of the particle set is around 60, which guarantees good accuracy while not costing too much in computation.
\begin{figure}[!ht]
  \centering
  \vspace*{-5pt}
  \begin{subfigure}{.5\linewidth}
    \centering
    \includegraphics[width=\textwidth]{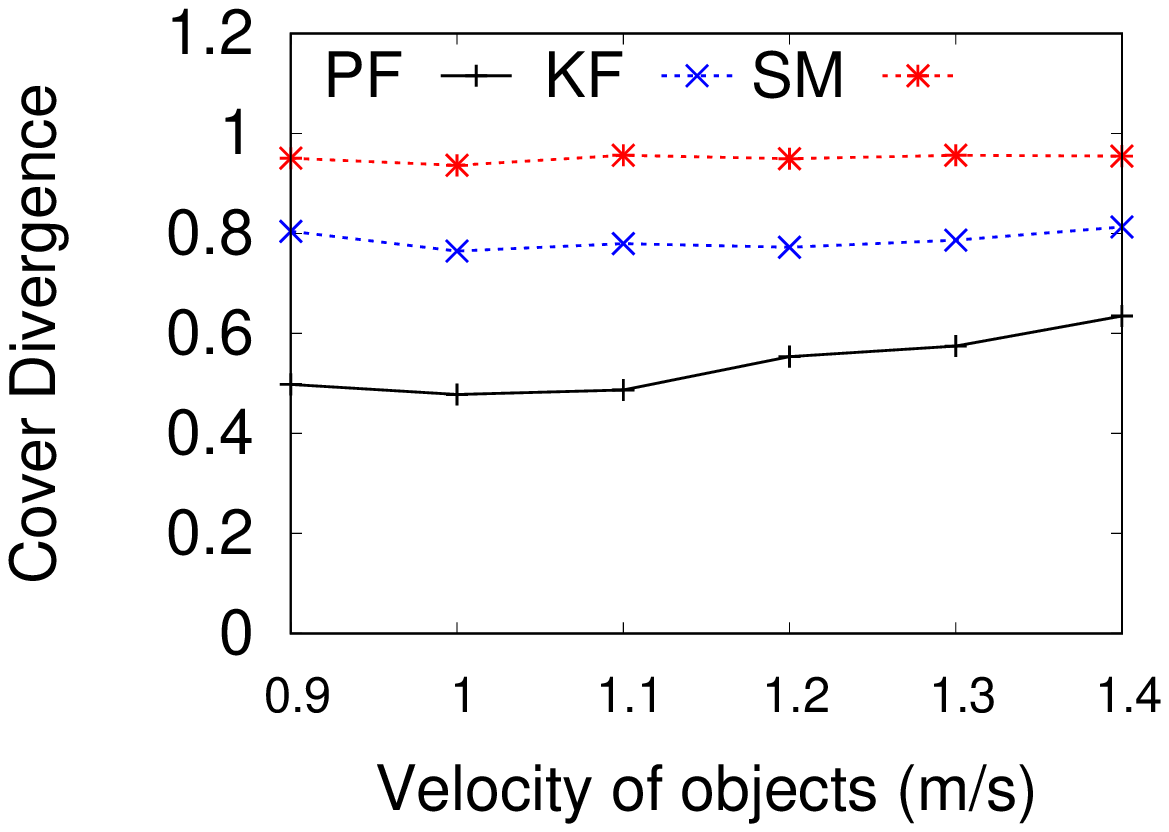}
    \caption{Cover divergence}
  \end{subfigure}%
  \begin{subfigure}{.5\linewidth}
    \centering
    \includegraphics[width=\textwidth]{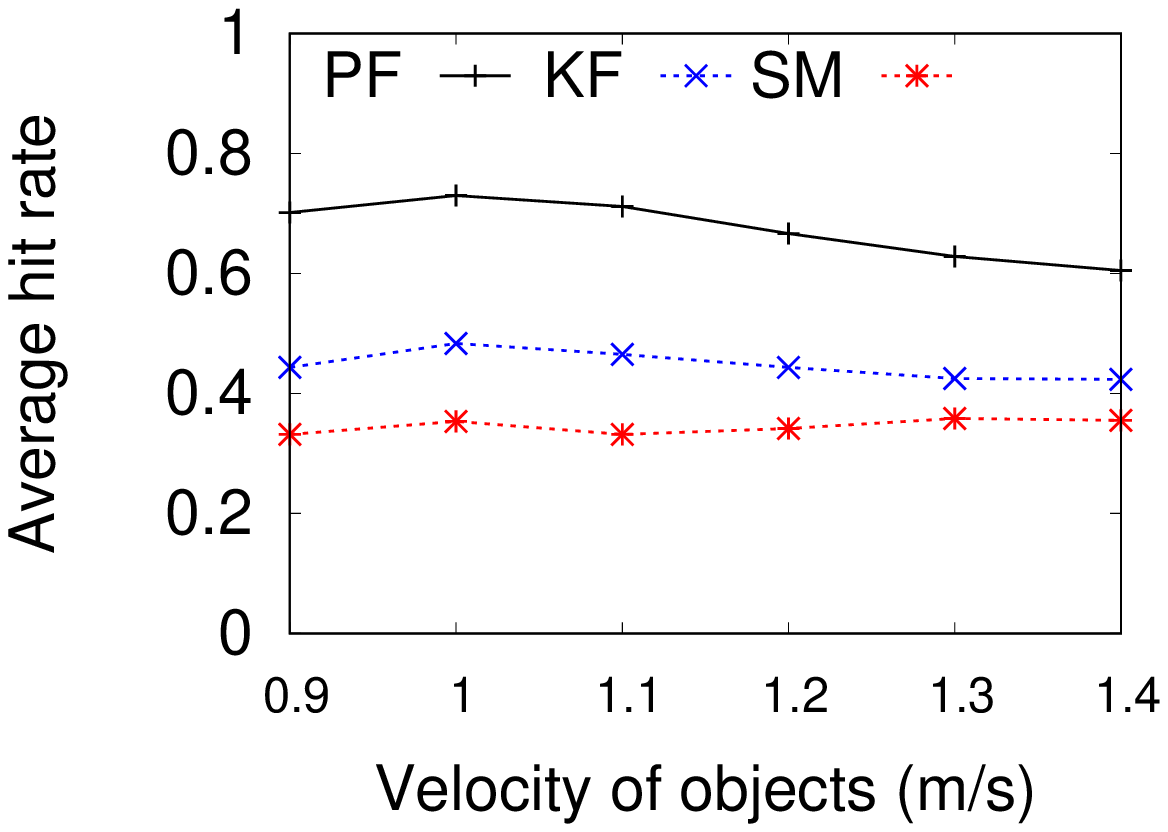}
    \caption{$k$NN success ratio}
  \end{subfigure}
  \vspace{-3mm}
  \caption{Result of varying the moving speed of objects.}
  \label{fig:vel}
  \vspace{-5mm}
\end{figure}

\subsubsection{Effects of Speed of Moving Objects}
To justify the assumption about velocity made in this paper, we generate the trajectories of objects with different velocities. In the experiment, we vary the constant moving speed \cite{Yang:indoorrange} of the objects from 0.9 m/s to 1.4 m/s to get the ground truth. Figure \ref{fig:vel} shows the performance of the three models. The PF model outperforms the other two models at all moving speed of objects. And the KF model exceeds SM. We get the same comparison result as that of the default experimental setting (a Gaussian distribution with
$\mu=1$ m/s and $\sigma=0.1$ m/s).

\subsubsection{Effects of Number of Moving Objects}

\begin{figure}[!ht]
  \centering
  \vspace{-5mm}
  \begin{subfigure}{.5\linewidth}
    \centering
    \includegraphics[width=\textwidth]{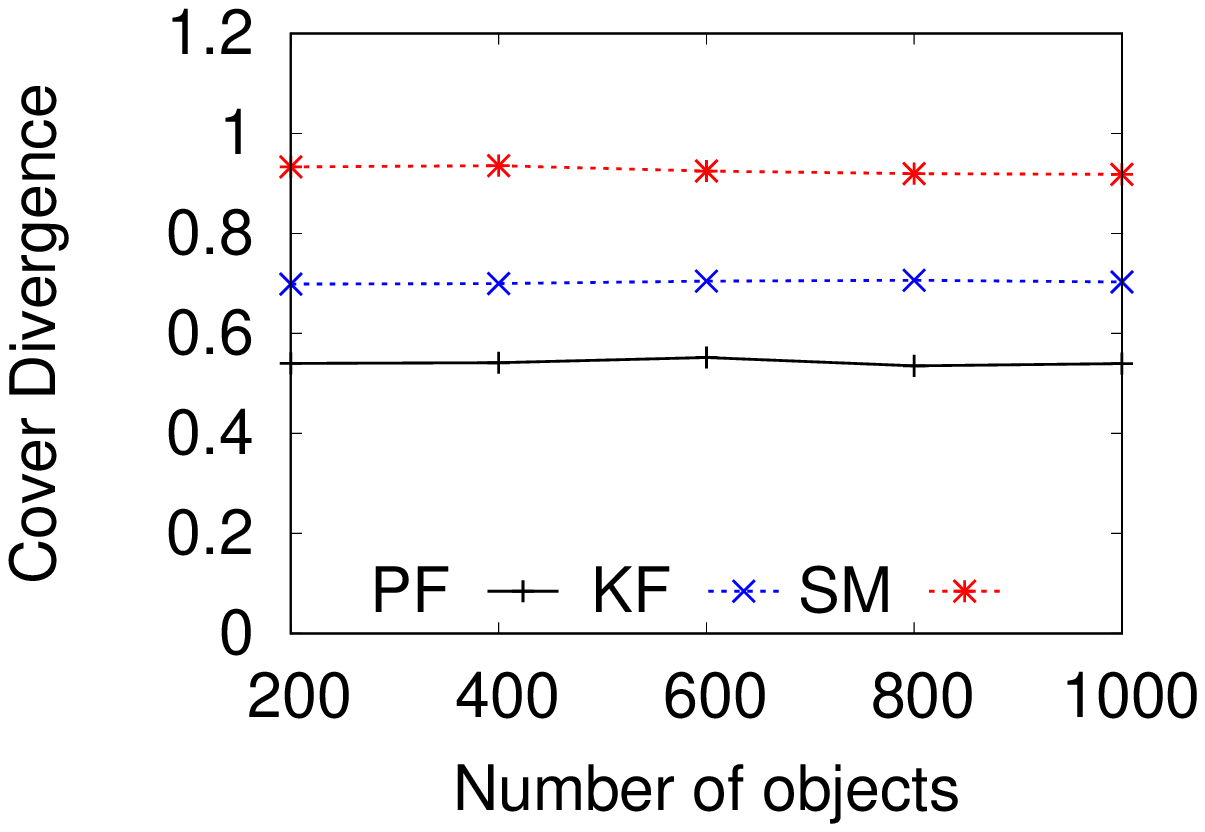}
    \caption{Cover divergence}
  \end{subfigure}%
  \begin{subfigure}{.5\linewidth}
    \centering
    \includegraphics[width=\textwidth]{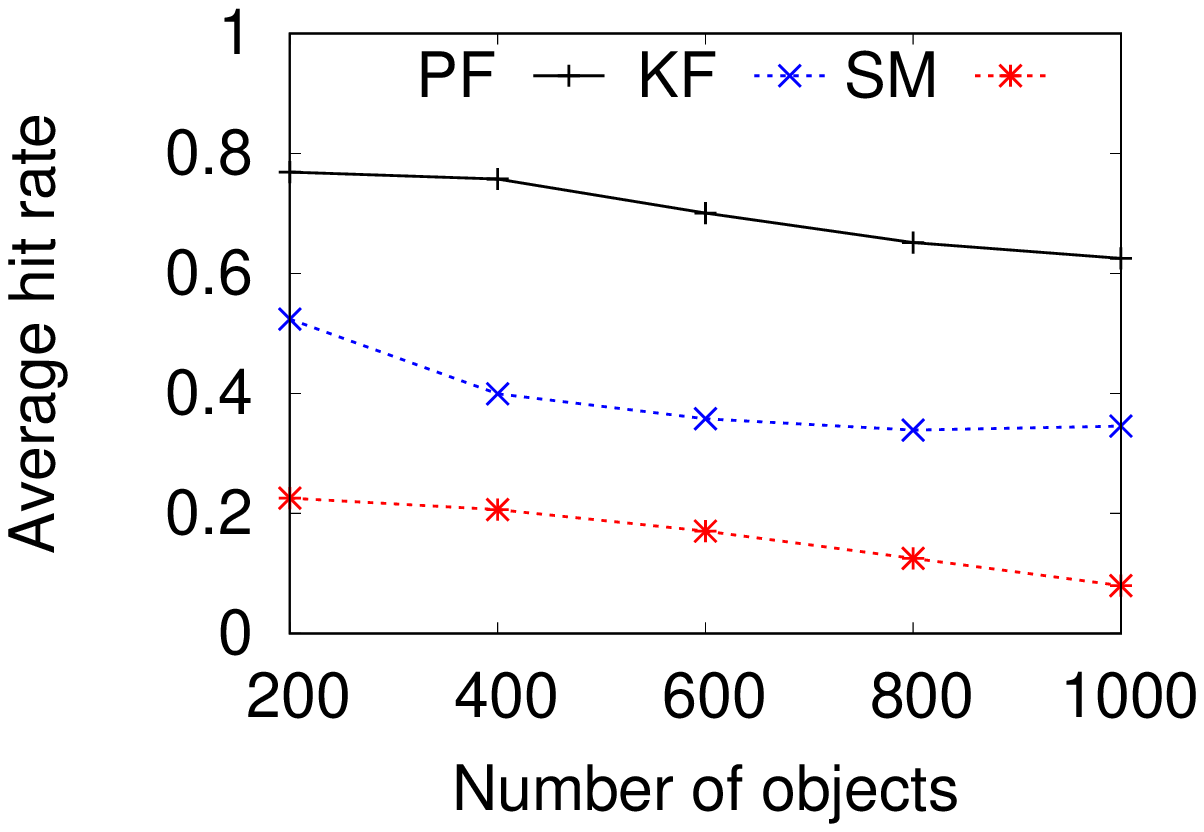}
    \caption{$k$NN success ratio}
  \end{subfigure}
  \vspace{-3mm} 
  \caption{The impact of the number of moving objects.}
  \vspace*{0pt}
  \label{fig:obj}
  \vspace{-5mm}
\end{figure}

In this subsection, we evaluate the scalability of our proposed
algorithms by varying the number of moving objects from 200 to
1000. All the result data are collected by averaging an extensive
number of queries over different query locations and time stamps.
Figure~\ref{fig:obj} shows that the cover divergence of the three
methods is relatively stable, while the average hit rate of $k$NN
queries decreases for all the methods. The decrease of $k$NN hit
rate is caused by increasing density of objects. A finer
resolution algorithm is required to accurately answer $k$NN
queries. In all, our solution demonstrates good scalability in
terms of accuracy when the number of objects increases.

\subsubsection{Effects of Activation Range}

\begin{figure}[!ht]
  \centering
  \begin{subfigure}{.5\linewidth}
    \centering
    \includegraphics[width=\textwidth]{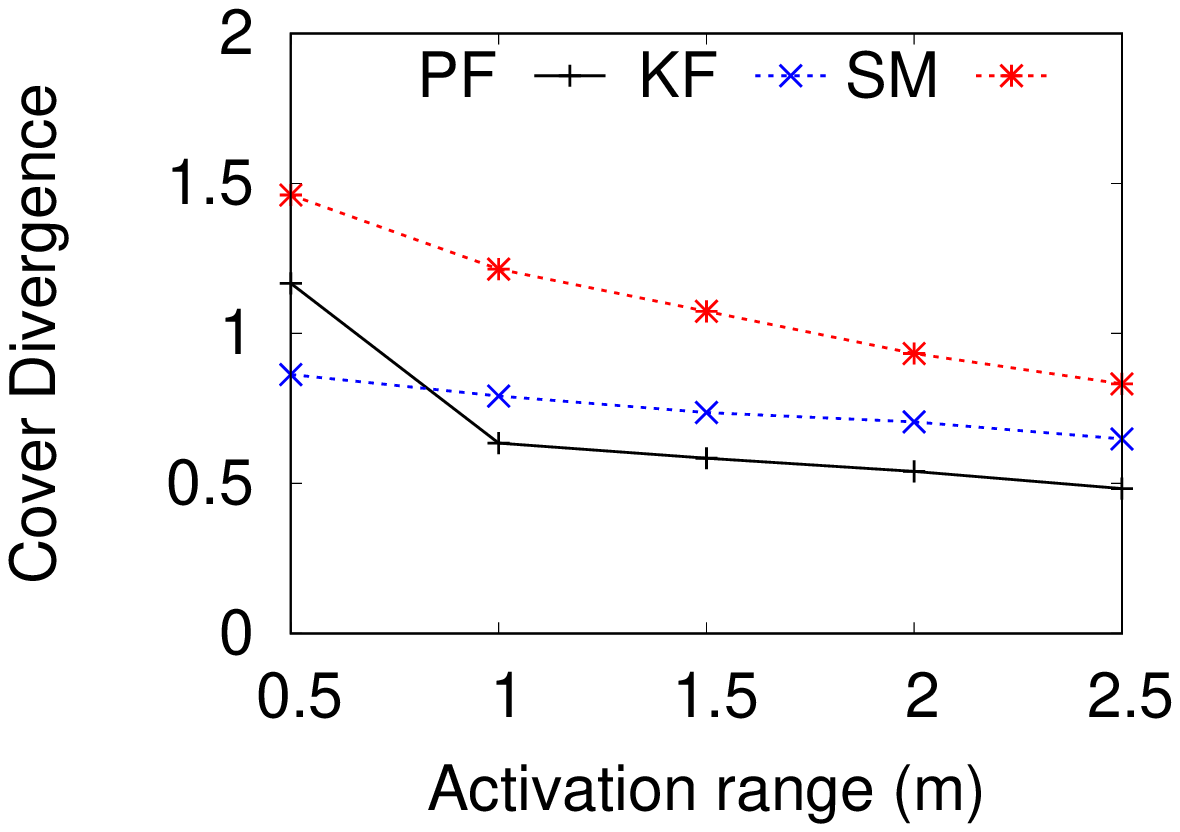}
    \caption{Cover divergence}
  \end{subfigure}%
  \begin{subfigure}{.5\linewidth}
    \centering
    \includegraphics[width=\textwidth]{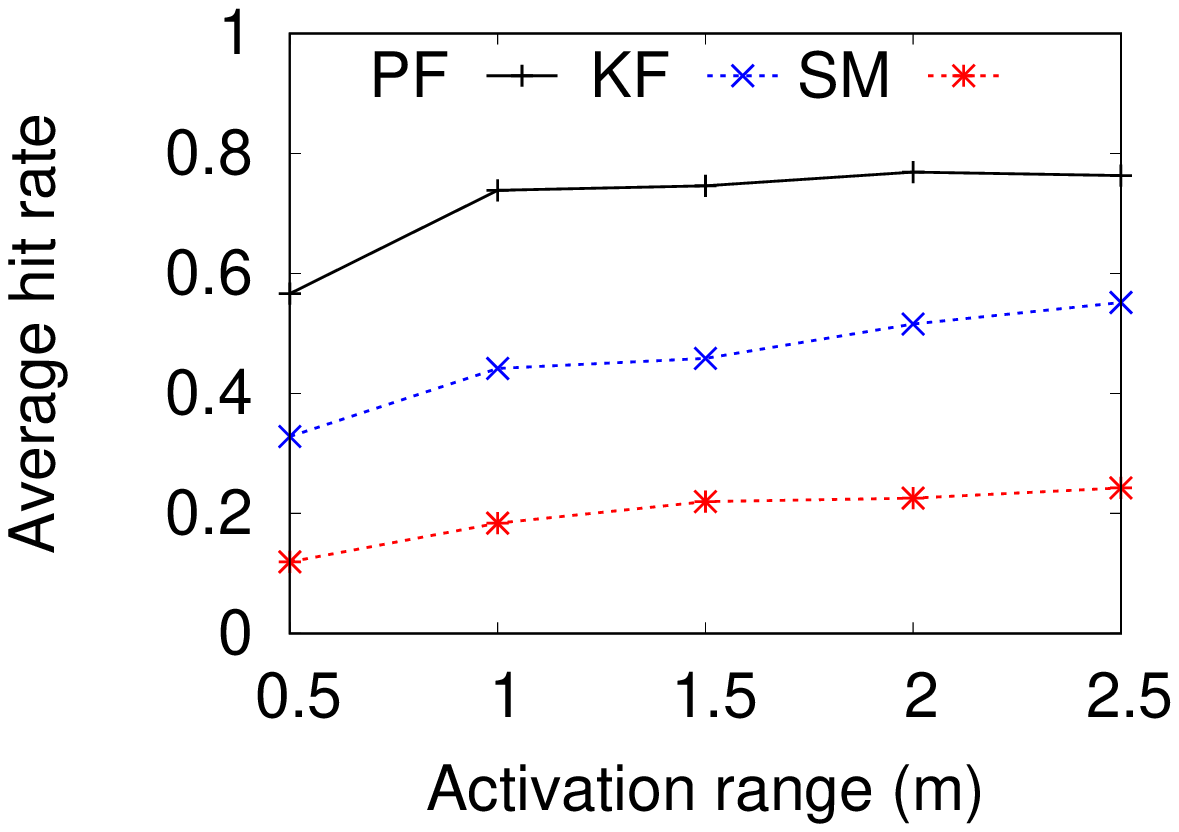}
    \caption{$k$NN success ratio}
  \end{subfigure}
  \vspace{-3mm}
  \caption{The impact of activation range.}
  \vspace*{-5pt}
  \label{fig:range}
\end{figure}

In this subsection, we evaluate the effects of the reader's
activation range by varying the range from 50 cm to 250 cm.  The
results are reported in Figure~\ref{fig:range}.  As the activation
range increases, the performance of all the three methods improves because uncertain regions not covered by any reader
essentially get reduced. In addition, even when the activation
range is small (e.g., 100 cm), the particle filter-based method is
still able to achieve relatively high accuracy.  Therefore, the
particle filter-based method is more suitable than the other two
methods when the physical constraints limit readers' activation
ranges.
\subsubsection{Continuous Query Performance Evaluation}

The previous subsections show the performance of snapshot queries,
i.e., queries at a specific time stamp.  This subsection
demonstrates our algorithms' performance across a duration of
time.  The application scenarios are described as follows:
\begin{figure}[ht]
  \centering
  \vspace{-5mm}
  \includegraphics[width=3.0in]{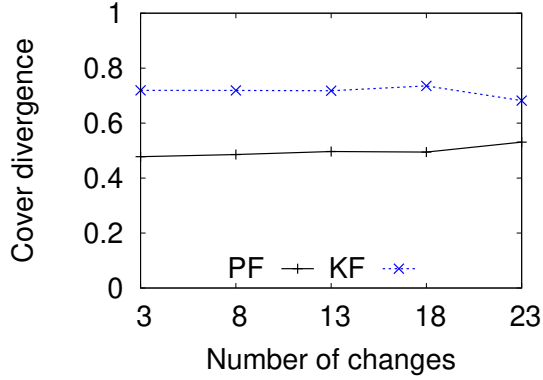}
  \vspace{-3mm}
  \caption{The impact of the number of changes.}
  \vspace*{-10pt}
  \label{fig:cont_n}
  \vspace{-2mm}
\end{figure}
\begin{enumerate}
\item For continuous range queries, a user registers a query window
at time $t_0$, and unregisters at $t_1$. During the time interval
(between $t_0$ and $t_1$), we keep updating the user of the
objects in the query window whenever a change is detected.

\item For continuous $k$NN queries, a user registers a query point
$q$ on the walking graph (a query point which is not on the
walking graph can be projected to its closest edge of the graph)
at $t_0$, and unregisters at $t_1$. During the time interval,
every time there is a change in the $k$ nearest neighbor query
result set, we will update the user with the new query result.
\end{enumerate}
\begin{figure}[!ht]
  \centering
  \begin{subfigure}{.5\linewidth}
    \centering
    \includegraphics[width=\textwidth]{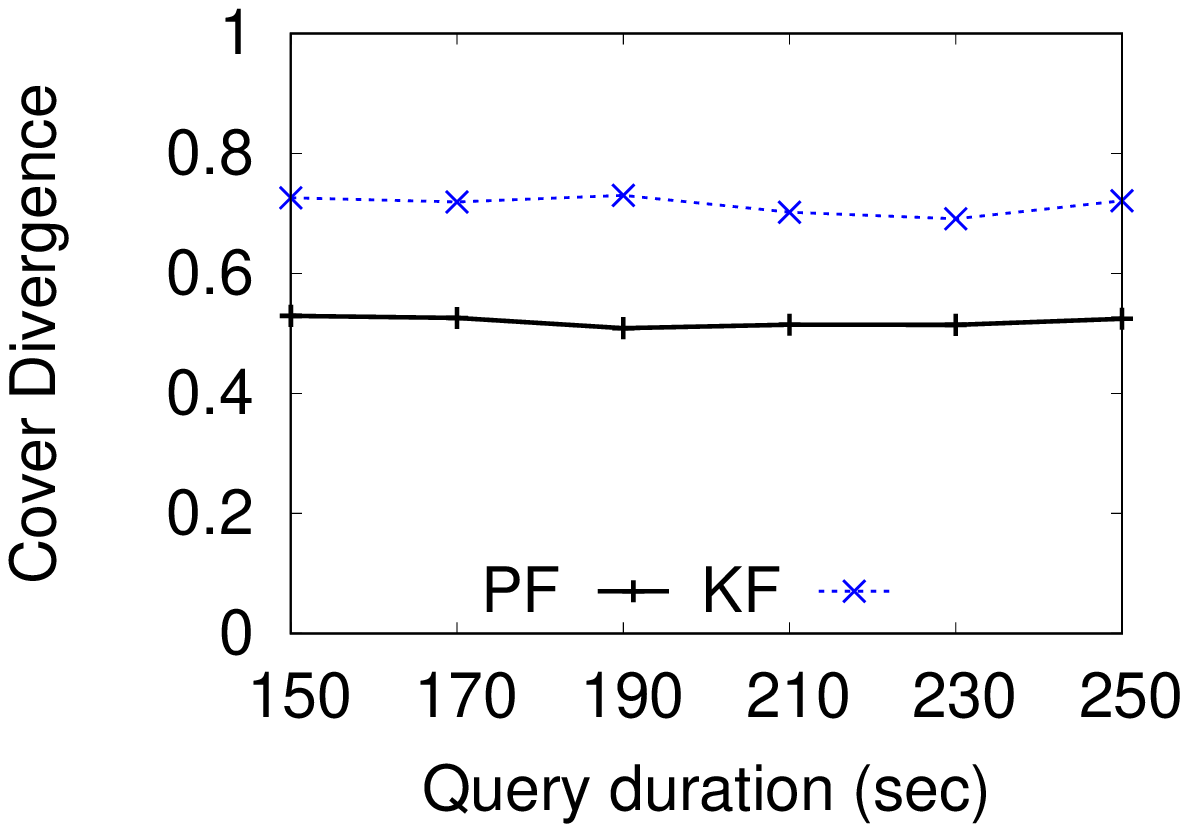}
    \caption{Continuous range query}
  \end{subfigure}%
  \begin{subfigure}{.5\linewidth}
    \centering
    \includegraphics[width=\textwidth]{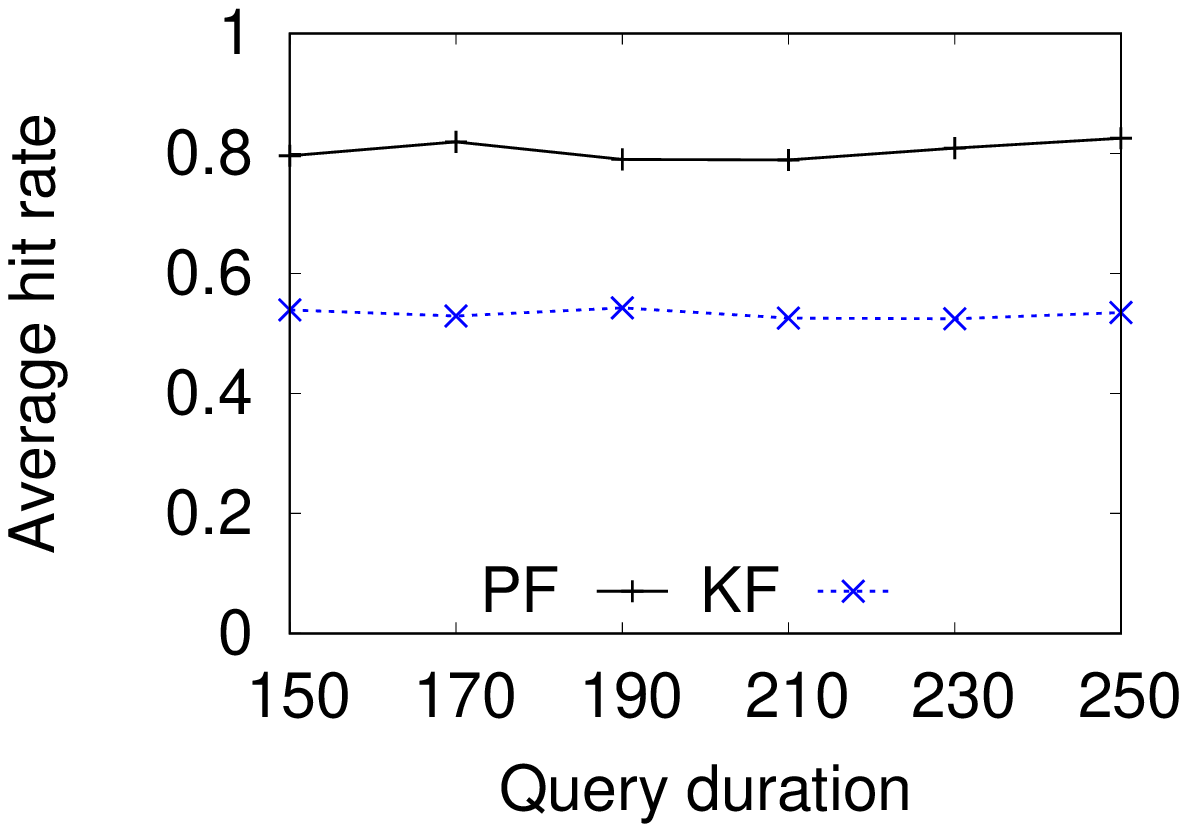}
    \caption{Continuous $k$NN query}
  \end{subfigure}
  \vspace{-1mm}
  \caption{The impact of query duration.}
  \vspace*{-5pt}
  \label{fig:cont_t}
\end{figure}
We develop two criteria to measure the performance in the above
scenarios:

\noindent \emph{\textbf{Change Volume}}: Change volume is defined as the
number of changes of objects in the query range between two
consecutive time stamps, including departing and arriving objects.
Suppose at $t_0$, the objects in the query range are $\{a, b,
c\}$; at $t_1$, the result set changes to $\{a, b, d\}$, then the
number of changes equals to 2, because one of the objects, $c$, is
departing and another object, $d$, just arrived. The rationale
behind this is that higher change volume could potentially impair
query result accuracy.

\noindent \emph{\textbf{Query Duration}}: Query duration is the interval between $t_0$ and $t_1$, where $t_0$ denotes the time a user
registers a continuous query, and $t_1$ denotes the time a user
unregisters the query. The rationale for this criteria is that the
proposed algorithms can be evaluated as stable and reliable if
they can maintain a satisfactory accuracy for a long duration.
Figure~\ref{fig:cont_n} shows the performance of our proposed
algorithms with different number of changes. It is clear from the
figure that our algorithms' accuracy is not heavily influenced by
the change volume, although there are some fluctuations. Updating the user of the objects in the query window once a change is detected contributes to the stability of performance.

Furthermore, Figure~\ref{fig:cont_t} shows the accuracy of our
algorithms against the query duration. Once the system is stable,
the accuracy of our algorithms is not affected by the duration of
query time.

\subsection{Real Data Set}
In the experiments utilizing real data, 40 objects were randomly moving on the second floor of the Haley Center on Auburn University campus; the trajectories were recorded by a camera. The experiments assumed that the RF readers were located at the designated positions. Once the object on the trajectory enters into the detection range of readers, it will be recorded with a specific probability and the hash table \texttt{AptoObjHT} will be updated. We evaluate all three models (PF, KF, and SM) with the collected data. 

\begin{figure}[ht]
  \centering
  \begin{minipage}{.5\linewidth}
      \includegraphics[width=\textwidth]{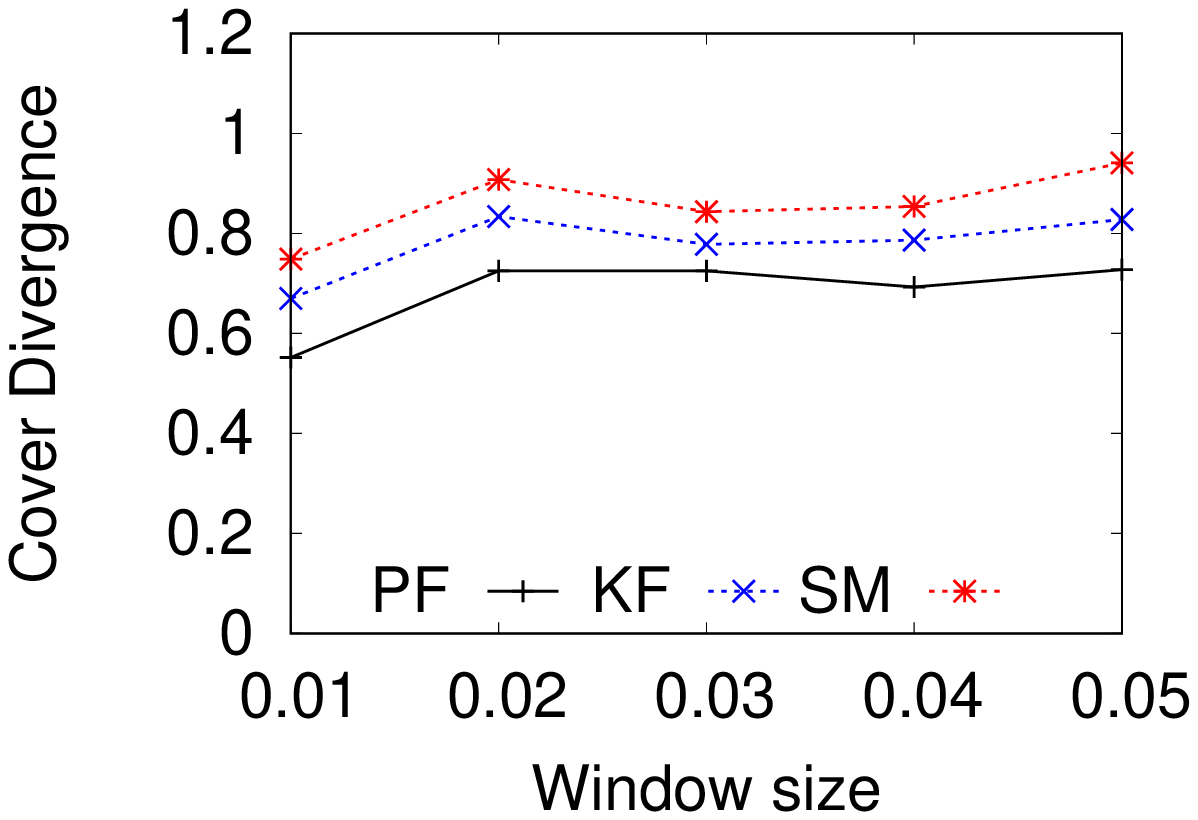}
      \caption{Effects of query window size.}
      \label{fig:real_window}
  \end{minipage}%
  \begin{minipage}{.5\linewidth}
      \includegraphics[width=\textwidth]{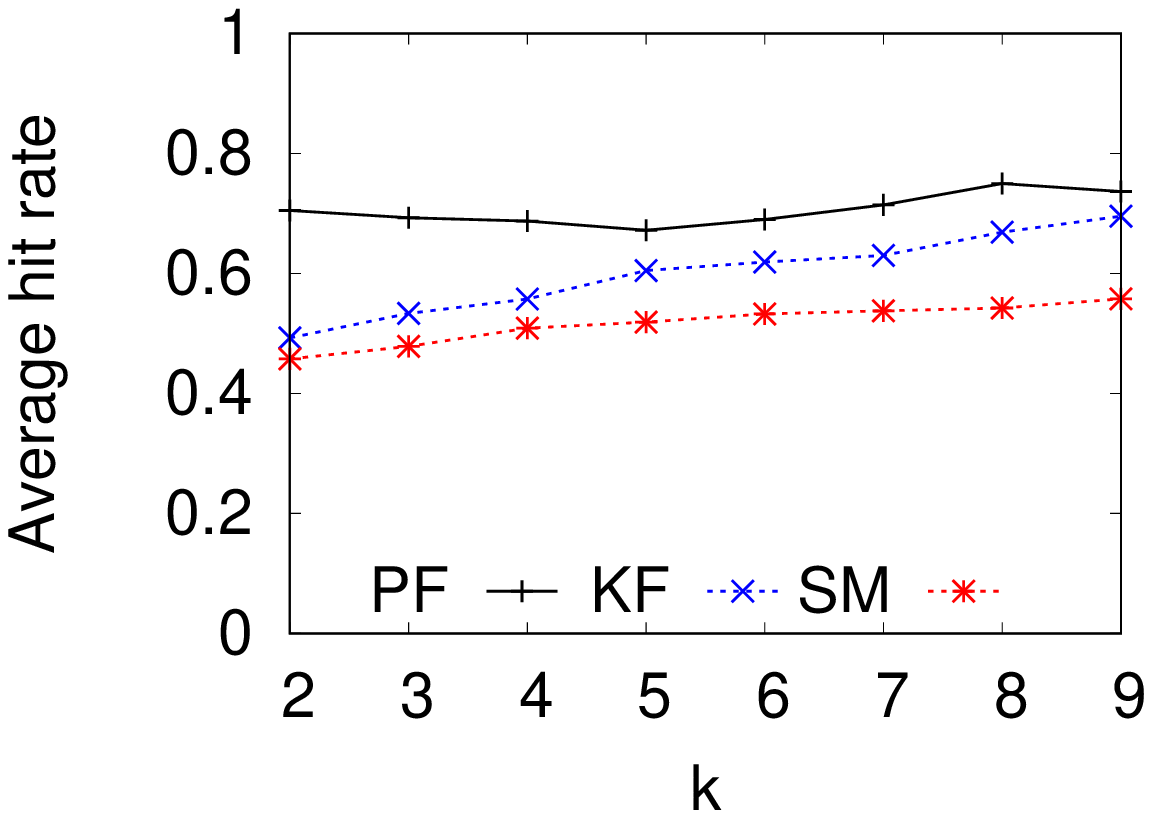}
      \caption{Effects of $k$.}
      \label{fig:real_hit}
  \end{minipage}%
\end{figure}  

\begin{figure}[ht]
  \centering
  \vspace*{-5pt}
  \begin{subfigure}{.5\linewidth}
    \centering
    \includegraphics[width=\textwidth]{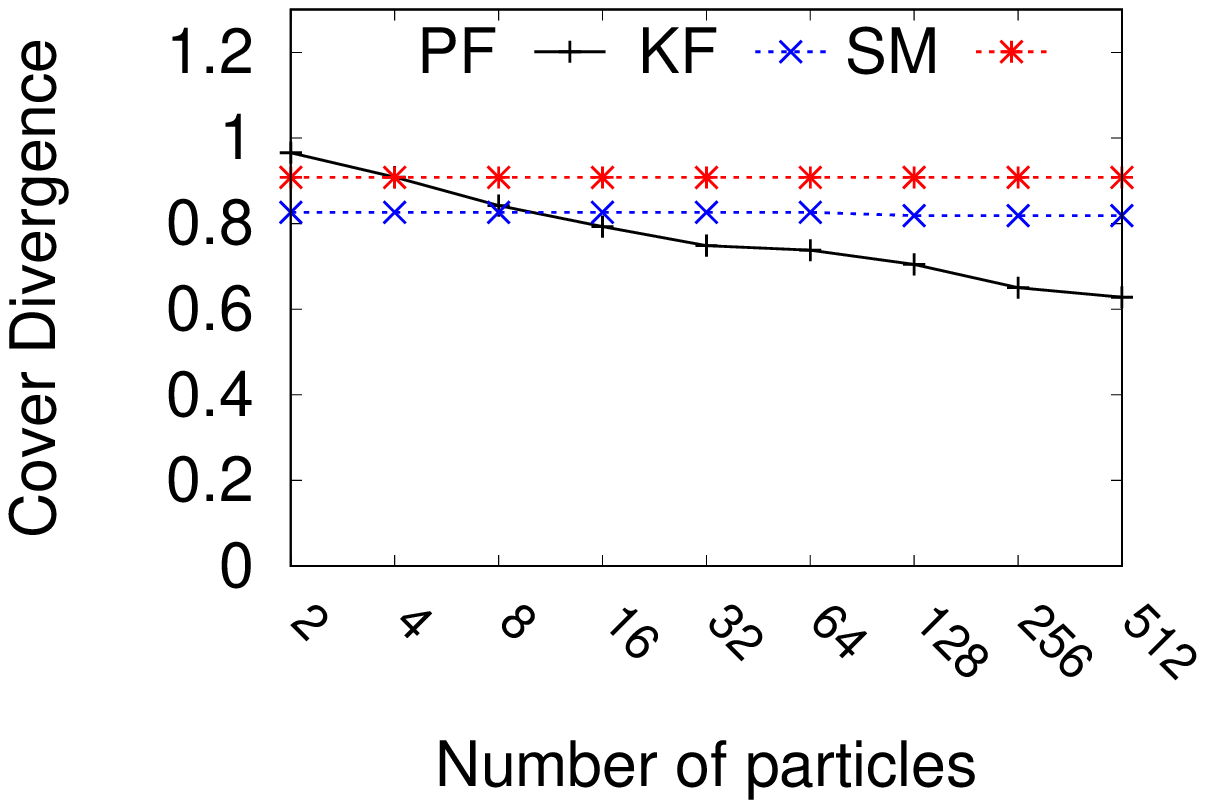}
    \caption{Cover divergence}
  \end{subfigure}%
  \begin{subfigure}{.5\linewidth}
    \centering
    \includegraphics[width=\textwidth]{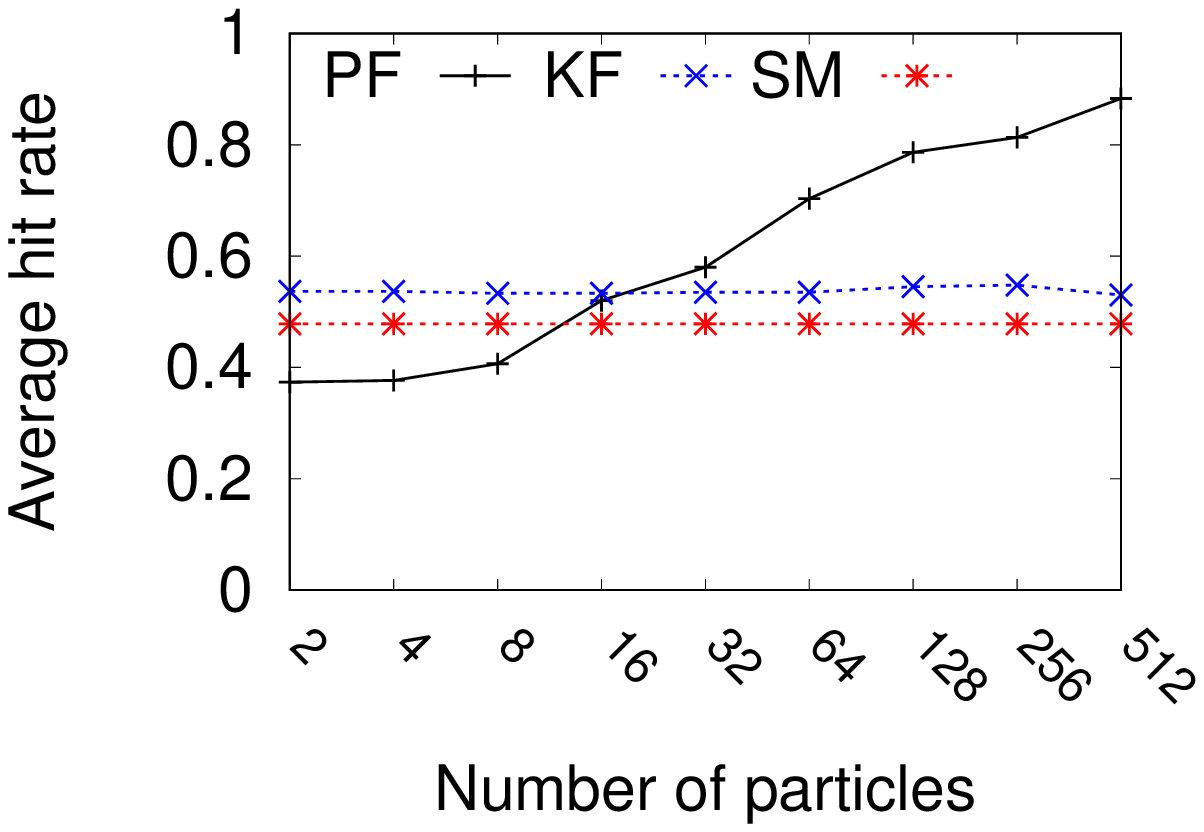}
    \caption{$k$NN success ratio}
  \end{subfigure}
  \vspace{-3mm}
  \caption{The impact of the number of particles.}
  \label{fig:real}
  \vspace{-5mm}
\end{figure}

\vspace{-1mm}
 Figure~\ref{fig:real_window} shows the effects of the query window size. The result is not significantly influenced by the query window size when the window size is greater than 0.01. When the query window size is 0.01, the query window cannot cover the whole room or the width of the hallway. At the same time, the number of moving objects is small. As a result, the cover divergence is relatively small. As shown in Figure~\ref{fig:real_hit}, the hit rate of PF outperforms SM and KF for different $k$ values. As $k$ goes from 2 to 9, the average hit rates of KF and SM grow slowly. The hit rate of PF is stable relatively concerning the value of $k$. Figure~\ref{fig:real} shows the effects of varying the number of particles on the query result. As the number of particles grows beyond 16, the performance of PF exceeds the other two. The reason is that as the number of particles increases, more possible anchors could be the position of the specific object. As a result, the algorithm will return more objects. Since there is no particle in KF and SM, the result of KF and SM will not be influenced by the number of particles. Overall, the comparison result on the real data set is the same as that on the synthetic data set.


\section{Conclusion}\label{sec:conc}
In this paper, we introduced an RFID and Bayesian filtering-based
indoor spatial query evaluation system. In order to evaluate
indoor spatial queries with unreliable data collected by RFID
readers, we proposed the Bayesian filtering-based location
inference method, the indoor walking graph model, and the anchor
point indexing model for cleansing noisy RFID raw data. After the
data cleansing process, indoor range and $k$NN queries can be
evaluated efficiently and effectively by our algorithms. We conduct comprehensive experiments using both synthetic and real-world data. The results demonstrate that our solution outperforms the symbolic model-based method significantly in query result accuracy with the assumption that objects move at a constant rate of 1 m/s, without stopping, waiting, or making detours.

For future work, we plan to conduct further analyses of our system
with more performance evaluation metrics and object moving trajectory patterns (e.g., people may stop for a while at a certain location as in a shopping mall setting). In addition, we intend to extend our framework to support more spatial query types such as spatial skyline, spatial joins and closest-pairs.

\section{Acknowledgement}
This research has been funded in part by the U.S. National Science Foundation grants IIS-1618669 (III) and ACI-1642133 (CICI).

\bibliographystyle{ACM-Reference-Format}    
\bibliography{geo.bib}   

\end{document}